\documentclass[11pt]{article}

\usepackage[final]{acl}

\usepackage{times}
\usepackage{latexsym}
\usepackage[T1]{fontenc}
\usepackage[utf8]{inputenc}
\usepackage{microtype}
\usepackage{graphicx}
\usepackage{amsmath,amssymb}
\usepackage{booktabs}
\usepackage{multirow}
\usepackage{array}
\usepackage{xcolor}
\usepackage{tikz}
\usepackage{tabularx}
\newcolumntype{Y}{>{\raggedright\arraybackslash}X}
\usetikzlibrary{arrows.meta,positioning,calc,shapes.misc,fit,backgrounds}
\usepackage[edges]{forest}

\usepackage{xcolor}
\usepackage{tikz}
\usetikzlibrary{calc,positioning}
\usepackage{xurl}

\definecolor{LTTgoldFill}{HTML}{FFE6CC}
\definecolor{LTTgoldEdge}{HTML}{D79B00}

\definecolor{LTTgreenFill}{HTML}{D5E8D4}
\definecolor{LTTgreenEdge}{HTML}{82B366}

\definecolor{LTTpurpleFill}{HTML}{E1D5E7}
\definecolor{LTTpurpleEdge}{HTML}{9673A6}

\definecolor{LTTink}{HTML}{202020}
\definecolor{LTTline}{HTML}{666666}
\definecolor{LTTlink}{HTML}{245DB1}

\DeclareRobustCommand{\LTTsection}[1]{%
  \hyperref[#1]{\S\ref*{#1}}%
}

\definecolor{axA}{HTML}{DCE6F2}
\definecolor{axB}{HTML}{E2F0D9}
\definecolor{axC}{HTML}{FBE5D6}
\definecolor{axD}{HTML}{EDE1F3}
\definecolor{axE}{HTML}{FFF2CC}
\definecolor{axF}{HTML}{E0E0E0}
\definecolor{axAd}{HTML}{2F5597}
\definecolor{axBd}{HTML}{548235}
\definecolor{axCd}{HTML}{C55A11}
\definecolor{axDd}{HTML}{7030A0}
\definecolor{axEd}{HTML}{BF9000}
\definecolor{axFd}{HTML}{595959}
\definecolor{lane1}{HTML}{2F5597}
\definecolor{lane2}{HTML}{548235}
\definecolor{lane3}{HTML}{C55A11}

\usepackage[most]{tcolorbox}
\usepackage{enumitem}

\newtcolorbox{takeawaybox}{
    enhanced,
    width=\linewidth,
    colback=black!2,
    colframe=black!55,
    boxrule=0.5pt,
    arc=1.5mm,
    boxsep=0pt,
    left=6pt,
    right=6pt,
    top=5pt,
    bottom=5pt,
    before skip=10pt,
    after skip=10pt,
    drop shadow=black!18
}

\title{Looped Transformers: Architecture, Training, and Inference}

\author{
    \textbf{Minh Le}$^{1,}$\thanks{Equal contribution.} \quad
    \textbf{Truong Giang Doan}$^{2,*}$ \quad
    \textbf{Bao-Ngoc Dao}$^{3,*}$ \quad
    \textbf{Minh Chu Xuan}$^{2,*}$ \\ [0.1em]
    \textbf{Binh Anh Nguyen}$^{2,3,*}$ \quad
    \textbf{Pham Chau Long}$^{2,*}$ \quad
    \textbf{Huy Huu Nguyen}$^{5,*}$ \quad
    \textbf{Quyen Tran}$^{6}$ \\ [0.1em]
    \textbf{Quan Dao}$^{6}$ \quad
    \textbf{Yong Liu}$^{1}$ \quad
    \textbf{Dimitris N. Metaxas}$^{6}$ \\ [0.1em]
    \textbf{Jieyang Chen}$^{5}$ \quad
    \textbf{Yingyan (Celine) Lin}$^{1}$ \quad
    \textbf{Linh Ngo Van}$^{2}$ \\[0.4em]
    $^{1}$Georgia Institute of Technology \\ [0.1em]
    $^{2}$Hanoi University of Science and Technology \\ [0.1em]
    $^{3}$Center for AI Research, VinUniversity \\ [0.1em]
    $^{5}$University of Oregon \
    $^{6}$Rutgers University
}

\begin{document}
\maketitle

\begin{abstract}
Looped Transformers repeatedly apply shared layers or blocks to evolving hidden representations, enabling deeper computation without a proportional increase in model size and, in some architectures, flexible inference depths. Despite growing interest in looped Transformers, understanding when recurrent computation improves model quality or computational efficiency remains challenging due to the diversity of design choices and evaluation settings. In this paper, we present a unified analysis of looped Transformers for language modeling and NLP along three complementary design axes: architecture, training, and inference. Beyond organizing existing approaches, we synthesize theoretical and empirical evidence to identify common patterns, characterize trade-offs, and examine the conditions under which recurrent computation is beneficial. Our analysis highlights that the benefits of parameter sharing depend on resource constraints and evaluation metrics, while reductions in recurrent computation do not necessarily translate into practical efficiency gains. Building on these findings, we propose a framework for selecting and evaluating looped Transformers under application-specific constraints and illustrate its use through two deployment scenarios. We further identify open challenges and future research directions. The full list of articles examined in this work is available at \url{https://github.com/ngocdaobao/Awesome-Looped-Transformers}.
\end{abstract}


\newcommand{\dbzijn}{\Delta \beta_{0ij}^{n}}
\newcommand{\dboijn}{\Delta \beta_{1ij}^{n}}
\newcommand{\daijn}{\Delta a_{ij}^{n}}
\newcommand{\dbijn}{\Delta b_{ij}^{n}}
\newcommand{\dsijn}{\Delta \sigma_{ij}^{n}}

\newcommand{\bzin}{\beta^{n}_{0i}}
\newcommand{\boin}{\beta^{n}_{1i}}
\newcommand{\ain}{a_i^n}
\newcommand{\bin}{b_i^n}
\newcommand{\sigmain}{\sigma_i^n}

\newcommand{\bzj}{\beta_{0j}^{*}}
\newcommand{\boj}{\beta_{1j}^{*}}
\newcommand{\aj}{a_{j}^{*}}
\newcommand{\bj}{b_{j}^{*}}
\newcommand{\sigmaj}{\sigma_{j}^{*}}

\newcommand{\bzjp}{\beta_{0j'}^{*}}
\newcommand{\bojp}{\beta_{1j'}^{*}}
\newcommand{\ajp}{a_{j'}^{*}}
\newcommand{\bjp}{b_{j'}^{*}}
\newcommand{\sigmajp}{\sigma_{j'}^{*}}

\newcommand{\zerod}{\mathbf{0}_d}
\newcommand{\ktilde}{\tilde{k}}
\newcommand{\Dtilde}{\widetilde{D}}

\newcommand{\brj}{\bar{r}(|\mathcal{A}_j|)}
\newcommand{\trj}{\tilde{r}(|\mathcal{A}_j|)}
\newcommand{\trjp}{\tilde{r}(|\mathcal{A}_{j'}|)}

\newcommand{\brone}{\bar{r}(|\mathcal{A}_1|)}
\newcommand{\dboione}{\Delta_{t_2} \beta_{1i1}^{n}}
\newcommand{\dbzione}{\Delta \beta_{0i1}^{n}}
\newcommand{\daione}{\Delta a_{i1}^{n}}
\newcommand{\dbione}{\Delta b_{i1}^{n}}
\newcommand{\dsione}{\Delta \sigma_{i1}^{n}}

\newcommand{\trone}{\tilde{r}(|\mathcal{A}_1|)}
\newcommand{\trs}{\tilde{r}(|\mathcal{A}_{j^*}|)}

\newcommand{\dint}{\mathrm{d}}

\newcommand{\brackets}[1]{\left[ #1 \right]}
\newcommand{\parenth}[1]{\left( #1 \right)}
\newcommand{\bigparenth}[1]{\big( #1 \big)}
\newcommand{\biggparenth}[1]{\bigg( #1 \bigg)}
\newcommand{\braces}[1]{\left\{ #1 \right \}}
\newcommand{\abss}[1]{\left| #1 \right |}
\newcommand{\angles}[1]{\left\langle #1 \right \rangle}
\newcommand{\tp}{^\top}
\def\st{\textnormal{s.t.}}
\def\sgn{\texttt{sign}}
\newcommand{\norm}[1]{\left\lVert#1\right\rVert}

\def\TM{\texttt{T}}
\def\OT{\textnormal{OT}}
\def\TW{\textnormal{TW}}
\def\TSW{\textnormal{TSW}}

\def\RR{\mathbb{R}}
\def\DD{\mathbb{D}}
\def\NN{\mathbb{N}}
\def\PP{\mathbb{P}}
\def\MM{\mathbb{M}}
\def\SS{\mathbb{S}}
\def\EE{\mathbb{E}}
\def\FF{\mathbb{F}}
\def\TT{\mathbb{T}}
\def\XX{\mathbb{X}}
\def\QQ{\mathbb{Q}}
\def\FF{\mathbb{F}}

\def\Ff{\mathcal{F}}
\def\Hh{\mathcal{H}}
\def\Gg{\mathcal{G}}
\def\Ee{\mathcal{E}}
\def\Pp{\mathcal{P}}
\def\Ss{\mathcal{S}}
\def\Ww{\mathcal{W}}
\def\Ff{\mathcal{F}}
\def\Rr{\mathcal{R}}
\def\Nn{\mathcal{N}}
\def\Xx{\mathcal{X}}
\def\Tt{\mathcal{T}}
\def\Mm{\mathcal{M}}
\def\Qq{\mathcal{Q}}

\newcommand{\Sbm}{{\bm S}}

\newcommand{\xbm}{{\bm x}}
\newcommand{\Xbm}{{\bm X}}
\newcommand{\Xbf}{{\mathbf{X}}}

\newcommand{\ybm}{{\bm y}}
\newcommand{\Ybm}{{\bm Y}}

\newcommand{\pbm}{{\bm p}}
\newcommand{\Pbm}{{\bm P}}
\newcommand{\Pbf}{{\mathbf{P}}}
\newcommand{\Gbf}{{\mathbf{G}}}
\newcommand{\Fbf}{{\mathbf{F}}}

\newcommand{\wbm}{{\bm w}}
\newcommand{\Wbm}{\bm{W}}
\newcommand{\Wq}{\Wbm^{\rm q}}
\newcommand{\Wk}{\Wbm^{\rm k}}
\newcommand{\Wv}{\Wbm^{\rm v}}

\newcommand{\bbm}{{\bm b}}
\newcommand{\bq}{\bbm^{\rm q}}
\newcommand{\bk}{\bbm^{\rm k}}

\newcommand{\Abm}{{\bm A}}
\newcommand{\cbm}{{\bm c}}

\newcommand{\kbm}{{\bm k}}
\newcommand{\Kbm}{{\bm K}}

\newcommand{\ubm}{{\bm u}}
\newcommand{\Ubm}{{\bm U}}

\newcommand{\vbm}{{\bm v}}
\newcommand{\Vbm}{{\bm V}}

\newcommand{\qbm}{{\bm q}}
\newcommand{\Qbm}{{\bm Q}}

\newcommand{\abm}{{\bm \alpha}}
\newcommand{\sbm}{{\bm s}}
\newcommand{\gbm}{{\bm g}}
\newcommand{\hbm}{{\bm h}}

\newcommand{\ebm}{{\bm e}}
\newcommand{\zbm}{{\bm z}}

\newcommand{\tbm}{{\bm t}}
\newcommand{\Tbm}{{\bm T}}

\newcommand{\veta}{{\bm \eta}}

\newcommand{\LS}{\mathcal{LS}}
\newcommand{\NS}{\mathcal{NS}}
\newcommand{\CS}{\mathcal{CS}}
\newcommand{\NCS}{\mathcal{NCS}}
\newcommand{\CSb}{\mathcal{CS}\text{-b}}
\newcommand{\CSd}{\mathcal{CS}\text{-d}}
\newcommand{\CSs}{\mathcal{CS}\text{-s}}
\newcommand{\NCSb}{\mathcal{NCS}\text{-b}}
\newcommand{\NCSd}{\mathcal{NCS}\text{-d}}
\newcommand{\NCSs}{\mathcal{NCS}\text{-s}}
\newcommand{\CSW}{\text{CSW}}
\newcommand{\NCSW}{\text{NCSW}}
\newcommand{\NCSWb}{\mathcal{NCSW}\text{-b}}
\newcommand{\NCSWd}{\mathcal{NCSW}\text{-d}}
\newcommand{\NCSWs}{\mathcal{NCSW}\text{-s}}

\newcommand{\pop}{F}
\newcommand{\nop}{F_n}
\newcommand{\popNGD}{F^{\texttt{NGD}}}
\newcommand{\nopNGD}{F_n^{\texttt{NGD}}}

\newcommand{\xbf}{\mathbf{x}}
\newcommand{\ybf}{\mathbf{y}}
\newcommand{\hbf}{\mathbf{h}}
\newcommand{\wbf}{\mathbf{w}}
\newcommand{\bbf}{\mathbf{b}}

\newcommand*{\vertbar}{\rule[-1ex]{0.5pt}{2.5ex}}
\newcommand*{\horzbar}{\rule[.5ex]{2.5ex}{0.5pt}}

\newcommand{\NormGD}{NormGD}
\newcommand{\ds}{\displaystyle}

\newcommand{\argmax}{arg\,max}
\newcommand{\argmin}{arg\,min}
\newcommand{\bbP}{\mathbb{P}}
\newcommand{\bbE}{\mathbb{E}}
\newcommand{\var}{\mathrm{Var}}

\newcommand{\softmax}{\mathrm{softmax}}
\newcommand{\sigmoid}{\mathrm{sigmoid}}
\newcommand{\gelu}{\mathrm{GELU}}

\def\st{{\em s.t.~}}
\def\ie{{\em i.e.,~}}
\def\eg{{\em e.g.,~}}
\def\cf{{\em cf.,~}}
\def\ea{{\em et al.~}}
\newcommand{\iid}{i.i.d.}
\newcommand{\wrt}{w.r.t.}

\newcommand{\deijn}{\Delta \eta_{ij}^{n}}

\newcommand{\dboin}{\Delta \beta_{1i}^{n}}
\newcommand{\dbzin}{\Delta \beta_{0i}^{n}}

\newcommand{\dain}{\Delta a_{i}^{n}}
\newcommand{\dbin}{\Delta b_{i}^{n}}
\newcommand{\dein}{\Delta \eta_{i}^{n}}

\newcommand{\cin}{c_i^n}
\newcommand{\ein}{\eta_i^n}

\newcommand{\boonen}{\beta_{11}^n}
\newcommand{\bzonen}{\beta_{01}^n}
\newcommand{\aonen}{a_1^n}
\newcommand{\bonen}{b_1^n}
\newcommand{\eonen}{\eta_1^n}

\newcommand{\coj}{c_j^0}
\newcommand{\coi}{c_i^0}

\newcommand{\aaoj}{A_j^0}
\newcommand{\aaoi}{A_i^0}

\newcommand{\eoj}{\eta_j^0}
\newcommand{\eoi}{\eta_i^0}

\newcommand{\cj}{c_j^*}

\newcommand{\ej}{\eta_j^*}

\newcommand{\boi}{\beta_{1i}^*}
\newcommand{\bzi}{\beta_{0i}^*}
\newcommand{\ai}{a_i^*}
\newcommand{\bi}{b_i^*}
\newcommand{\ei}{\eta_i^*}

\newcommand{\boone}{\beta_{11}^*}
\newcommand{\bzone}{\beta_{01}^*}
\newcommand{\aone}{a_1^*}
\newcommand{\bone}{b_1^*}
\newcommand{\eone}{\eta_1^*}

\newcommand{\cjp}{c_{j'}^0}
\newcommand{\gjp}{\Gamma_{j'}^0}
\newcommand{\ejp}{\eta_{j'}^0}

\newcommand{\zeroq}{\mathbf{0}_q}
\newcommand{\pizeroone}{\pi_{1}^{0}}
\newcommand{\dtone}{\Delta \tau^{n}}

\newcommand{\deione}{\Delta \eta_{i1}^{n}}

\newcommand{\normf}[1]{\|#1\|_{L^2(\mu)}}

\newcommand{\bfit}[1]{\boldsymbol{#1}}

\newcommand{\prompt}{\bm p}
\newcommand{\dt}{\mathcal{D}_t}
\newcommand{\data}{\mathcal{D}}

\newcommand{\xdom}{\mathcal{X}^t}
\newcommand{\ydom}{\mathcal{Y}^t}

\newcommand{\yi}{\mathcal{Y}^{(i)}}
\newcommand{\yj}{\mathcal{Y}^{(j)}}

\newcommand{\normop}{\mathcal{S}_p}
\newcommand{\att}{\mathrm{Attention}}
\newcommand{\dv}{d_v}
\newcommand{\dk}{d_k}

\def\mmoe{\texttt{MMoE}}

\definecolor{indigo}{RGB}{75, 0, 130}          
\newcommand{\minh}[1]{\textcolor{indigo}{[Minh: #1]}}


\begin{figure*}[t]
\vspace{-0.7em}
    \centering
    \input{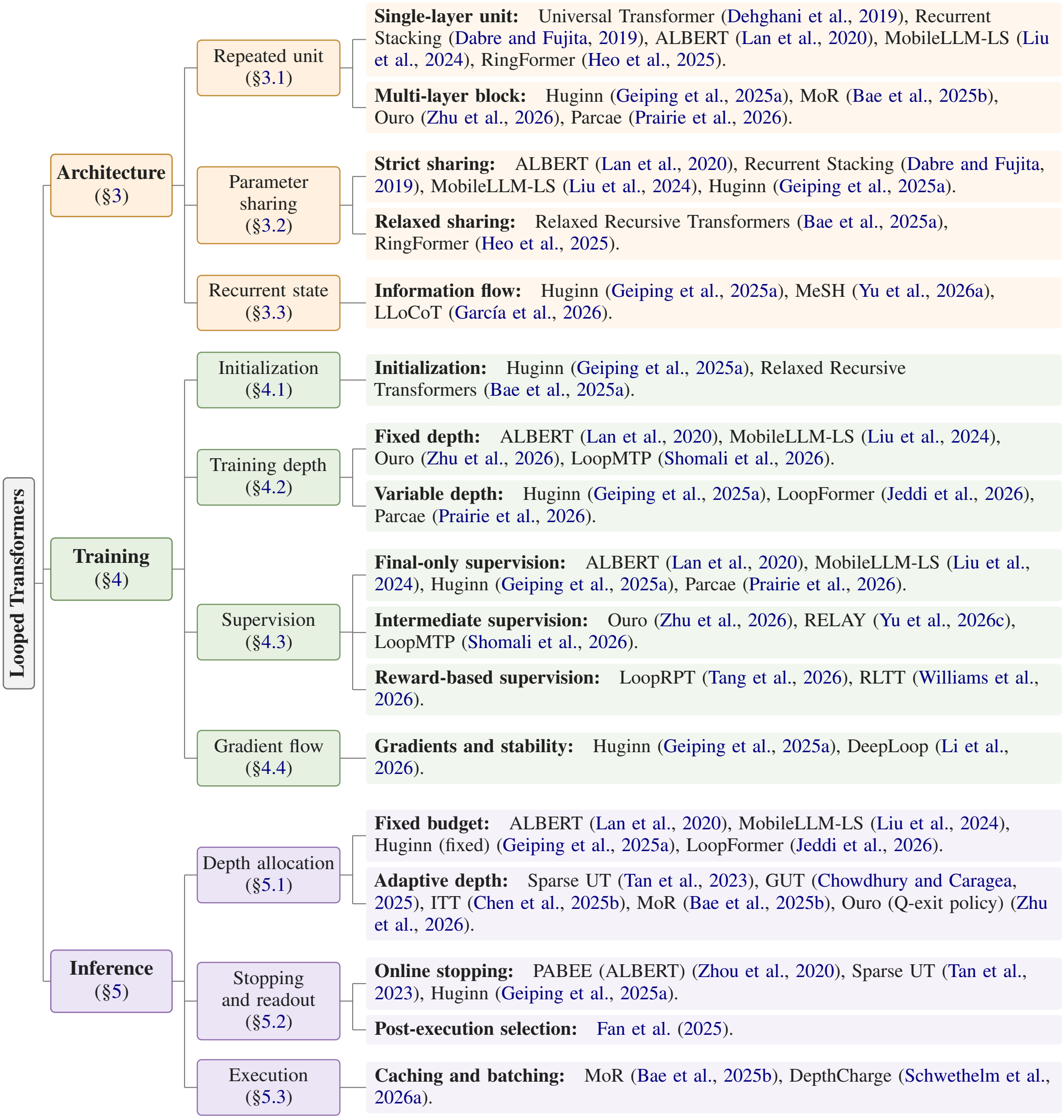}
    \caption{Taxonomy of looped Transformers along three complementary design axes: architecture, training, and inference. Each category lists representative works, and a given work may appear in multiple categories.}
    \label{fig:looped_taxonomy}
    \vspace{-0.7em}
\end{figure*}

\section{Introduction}
\label{sec:introduction}

Conventional Transformers increase computational depth by stacking independently parameterized layers~\citep{vaswani2017attention}, whereas looped Transformers repeatedly apply shared layers or blocks to evolving hidden representations. Early examples include the Universal Transformer, recurrent stacking, and ALBERT~\citep{dehghani2018universal,dabre2019recurrent,lan2019albert}. Throughout this paper, we use the terms \emph{looped Transformer} and \emph{recurrent-depth Transformer} interchangeably. By decoupling computational depth from the number of distinct layers, parameter reuse enables deeper computation without a proportional increase in model size and, in some architectures, flexible inference depths~\citep{reid2021subformer,bae2025relaxed,geiping2026scaling}.

Despite growing interest in looped Transformers, understanding when recurrent computation is beneficial remains challenging. Existing studies pursue different objectives, such as increasing computational depth under a fixed parameter budget or reducing the parameter count at a fixed computational depth. Findings obtained under one resource constraint may therefore not generalize to another. Even under the same experimental controls, different evaluation metrics can lead to different conclusions. For example, when comparing a looped Transformer that applies a 12-layer block twice with an untied 24-layer Transformer at matched training-token counts and executed depth, \citet{saunshi2025reasoning} report higher mathematical reasoning accuracy but higher validation perplexity for the looped model. These differences complicate the identification of general principles for designing and evaluating looped Transformers.

Another challenge arises from the interdependence of architectural design, training, and inference. While architecture determines which computations are repeated, training influences how effectively the model utilizes additional iterations and whether intermediate states yield useful predictions. Both properties affect the effectiveness of adaptive inference policies that determine when further computation is necessary. Even when these policies reduce the number of executed iterations, practical efficiency gains depend on cache management and execution scheduling~\citep{zhu2026ouro,bae2026mixture}. Consequently, the benefits of individual design choices cannot be fully understood without considering their interactions.

Although existing studies offer valuable insights into recurrent computation, translating these findings into practical guidance for model design, training, and deployment remains an open challenge. To address this gap, we provide a unified analysis of the literature along three complementary design axes: architecture, training, and inference (Figure~\ref{fig:looped_taxonomy}). Beyond organizing existing approaches, we synthesize theoretical and empirical evidence to identify common design principles, characterize trade-offs, and examine the conditions under which different approaches are beneficial. Through this analysis, we aim to guide researchers and practitioners in designing and evaluating looped Transformers under different performance requirements and resource constraints.

\paragraph{Relation to existing surveys.}
Several surveys examine recurrent computation within broader research areas. Surveys of latent reasoning cover computation along both network depth and token positions~\citep{zhu2025latentsurvey,chen2025latentsurvey}, while \citet{ji2026survey} examine test-time computation more broadly, including sampling and search strategies. The survey most closely related to our work is that of \citet{azizi2026looped}, who examine recurrent-depth Transformers across application domains and analyze empirical findings under different resource constraints.

Complementing these surveys, we focus specifically on recurrent-depth computation in language modeling and NLP, encompassing both fixed-depth parameter sharing and adaptive recurrent computation. Our distinguishing emphasis is on \emph{configuration-level evidence}: we interpret reported findings by jointly considering the architectural choices, training procedures, inference policies, and evaluation settings of each configuration. This perspective allows us to clarify what the existing evidence establishes and how it can inform model design and deployment. Accordingly, we organize our empirical analysis around three questions:
\begin{enumerate}[label=\textbf{Q\arabic*.}, leftmargin=*, nosep]
    \item How does prediction quality vary with recurrence depth at inference for a fixed checkpoint?
    \item How do parameter sharing and other recurrent design choices affect model quality under controlled parameter, training-token, or compute budgets?
    \item Under what conditions do recurrent computation and adaptive inference improve practical efficiency, as measured by latency, throughput, and memory usage?
\end{enumerate}

\paragraph{Contributions.}
Our contributions are threefold.
\textbf{(i) Unified taxonomy.}
We organize the design choices of looped Transformers along three complementary axes---architecture, training, and inference---and examine representative mechanisms and their associated trade-offs (Figure~\ref{fig:looped_taxonomy}; Sections~\ref{sec:architecture}--\ref{sec:inference}).
\textbf{(ii) Configuration-level evidence synthesis.}
We examine how experimental controls, resource constraints, and evaluation metrics shape the interpretation of empirical findings. Through within-study comparisons and baseline-relative accuracy--compute analyses, we identify common patterns and characterize trade-offs (Section~\ref{sec:capabilities}). We also discuss theoretical guarantees, their assumptions, and their implications for evaluating trained models (Appendix~\ref{appendix:theory}).
\textbf{(iii) Practical design and evaluation guidance.}
We propose a framework for selecting and evaluating looped Transformers under application-specific resource constraints (Section~\ref{subsec:design_guide} and Appendix~\ref{app:design_evaluation}). Through two deployment scenarios, we illustrate how empirical evidence informs architectural choices and inference-depth selection. We further identify open challenges and future research directions (Section~\ref{sec:future}).
\section{Definitions and Scope}
\label{sec:definition_scope}

\subsection{What Is a Looped Transformer?}
\label{subsec:definition}

A \emph{looped Transformer} obtains part or all of its computational depth by repeatedly applying shared Transformer layers or blocks to refine hidden representations. Given an input $\xbf$ and a recurrence count $R$, we express a basic looped Transformer as
\begin{equation}
\begin{aligned}
\hbf_0 &= \Pbf(\xbf),\\
\hbf_{r+1} &= \Fbf_\theta(\hbf_r),\quad r=0,\ldots,R-1,\\
\hat{\ybf} &= \Gbf(\hbf_R).
\end{aligned}
\label{eq:strict_recurrent_core}
\end{equation}
Here, $\Pbf$ initializes the hidden state, $\Fbf_\theta$ is the \emph{recurrent core}, and $\Gbf$ maps the final state to the prediction. Across iterations, the core parameters $\theta$ remain shared, while the hidden state $\hbf_r$ is updated. The non-recurrent boundary modules $\Pbf$ and $\Gbf$ are optional. Appendix~\ref{app:architecture_state_notation} extends this formulation to persistent input conditioning, multiple states or cores, and iteration-dependent signals.

\begin{figure}[t]
    \centering
    \includegraphics[width=0.97\linewidth]{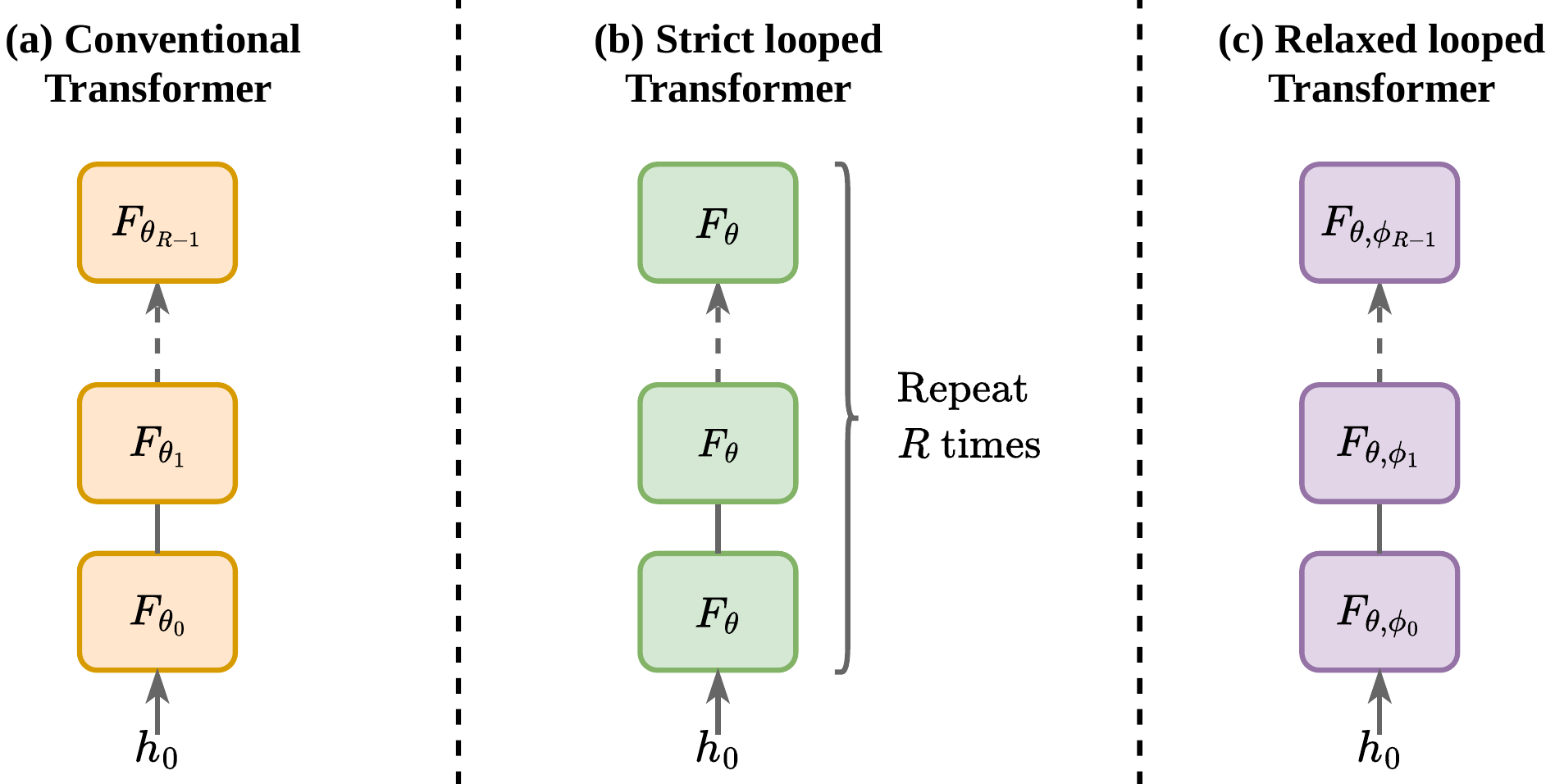}
\caption{\small Comparison of untied Transformers, which use distinct layer parameters, and looped Transformers with strict sharing (shared core parameters) or relaxed sharing (additional iteration-specific parameters).}
    \label{fig:looped_definition}
    \vspace{-1.0em}
\end{figure}

If the recurrent core consists of $L$ distinct layers, each applied once per iteration, we define $L$ as the \emph{unique core depth} and $RL$ as the \emph{effective recurrent depth}. Both exclude layers outside the recurrent core, and neither alone determines FLOPs or runtime. Equation~\eqref{eq:strict_recurrent_core} assumes \emph{strict sharing}, in which all core parameters are shared across iterations. In contrast, \emph{relaxed sharing} augments the shared parameters $\theta$ with iteration-specific parameters $\phi_r$, yielding $\hbf_{r+1}=\Fbf_{\theta,\phi_r}(\hbf_r)$, as exemplified by depth-specific low-rank corrections~\citep{bae2025relaxed}. Figure~\ref{fig:looped_definition} illustrates this distinction.

\subsection{Scope and Boundaries}

We include configurations in which shared Transformer layers or blocks repeatedly update hidden representations \emph{along network depth}. This criterion encompasses fixed-depth architectures with parameter sharing, such as ALBERT~\citep{lan2019albert}, without requiring adaptive stopping or depth extrapolation. Although PABEE applies a common early-exit policy to different backbones, its ALBERT configuration falls within our scope because of the backbone's parameter sharing rather than the early-exit mechanism itself~\citep{zhou2020pabee}.

Related mechanisms, such as segment recurrence~\citep{bulatov2022recurrent}, additional thought positions~\citep{hao2024training}, iterative denoising~\citep{nie2026large}, and skipping untied layers~\citep{raposo2024mixturedepths}, can increase or redistribute computation but do not, by themselves, constitute recurrent-depth computation. Hybrid architectures fall within our scope only if they contain an identifiable recurrent-depth component. For instance, LLoCoT qualifies because it repeatedly applies a shared block to refine a latent workspace before answer generation~\citep{graugarcia2026llocot}. Appendix~\ref{app:architecture_feedback} discusses additional boundary cases. Although architectures that employ these alternative mechanisms without recurrent-depth computation fall outside our scope, they can serve as relevant baselines for empirical comparison. 

\subsection{Review Procedure}
\label{subsec:review_organization}

We conducted a targeted narrative review of looped Transformers in language modeling and NLP, supplemented by selected theoretical, algorithmic, and technical studies that clarify the principles, capabilities, and limitations of recurrent-depth computation. Starting from the bibliography assembled during drafting, we conducted supplementary keyword searches and reference checks on 12 October 2026. Appendix~\ref{app:review_procedure} documents the search queries, the use of language-model suggestions to identify candidate studies, the selection criteria, and the limitations of the search process.

Figure~\ref{fig:looped_taxonomy} organizes \emph{design choices} along three complementary axes: architecture, training, and inference. For empirical analysis, we use the \emph{evaluated configuration} as the unit of analysis, with each configuration specified by its architecture, training procedure, inference policy, and evaluation setting. Rather than ranking method families across heterogeneous experimental settings, we report each result alongside its baseline, metric, and relevant resource controls.

\section{Architectural Design}
\label{sec:architecture}

The computation performed at each iteration of a looped Transformer depends on its recurrence structure, parameter-sharing scheme, and the information available to the recurrent core. We examine these design choices and their implications for recurrent computation. Appendix~\ref{appendix:architecture} provides additional notation and representative configurations.

\subsection{Recurrence Structure}
\label{subsec:recurrence_structure}

The recurrence structure determines which layers are repeated, in what order, and at what frequency. The composition of the recurrent core determines the computation performed at each iteration. For example, a single-layer Universal Transformer~\citep{dehghani2018universal} repeatedly applies one Transformer layer, whereas Huginn~\citep{geiping2026scaling} repeatedly applies a four-layer recurrent core while keeping its input and output layers outside the recurrence.

Even when the same layers are reused equally often, their execution order can lead to different hidden-state transformations~\citep{takase-kiyono-2023-lessons}. For example, the sequence $A,B,A,B$ alternates between layers $A$ and $B$, whereas $A,A,B,B$ applies each layer twice consecutively. Beyond layer ordering, hierarchical architectures~\citep{wang2025hrm,wang2026hrm} allow multiple recurrent cores to operate at different update frequencies, with several low-level updates occurring within a single high-level update. In such architectures, the recurrence count and effective recurrent depth must be characterized separately for each core.

\vspace{-0.2em}
\subsection{Parameter Sharing}
\label{subsec:architecture_sharing}
\vspace{-0.1em}

Parameter-sharing schemes differ in which components are shared and whether iteration-specific parameters are introduced. ALBERT~\citep{lan2019albert} investigates three strategies: sharing attention layers, feed-forward layers, or both. Relaxed Recursive Transformers~\citep{bae2025relaxed} extend strict parameter sharing by introducing depth-specific low-rank corrections to shared weight matrices, enabling iteration-specific transformations at the cost of additional parameters.

Although parameter sharing can improve parameter efficiency, its effect on model quality varies across evaluation metrics. In the comparison discussed in Section~\ref{sec:introduction}, both models are trained on 250 billion tokens and execute 24 layer applications. The looped model, which applies a 12-layer block twice, achieves a higher average score across mathematical word-problem tasks than the untied model (34.3 vs.\ 29.3) but exhibits higher validation perplexity (7.90 vs.\ 7.40)~\citep{saunshi2025reasoning}. This comparison illustrates that improvements in downstream reasoning do not necessarily translate into improvements in language modeling.

Beyond the choice of evaluation metric, the effectiveness of parameter sharing also depends on its architectural objective: increasing computational depth at a fixed parameter count or reducing the parameter count at a fixed computational depth. RINS~\citep{alabdulmohsin2025rins} follows the former approach by repeatedly applying an early block, achieving lower validation loss at larger compute budgets under matched forward-pass FLOPs. In contrast, \citet{schwethelm2026recurrence} investigate the latter approach by replacing distinct layers with shared ones and report higher validation loss than the untied baseline with four recurrent iterations.

\begin{takeawaybox}
\raggedright
\textbf{Takeaway.}
\emph{The benefits of parameter sharing at fixed parameter count do not necessarily extend to fixed computational depth.}
\end{takeawaybox}

\subsection{State Updates and Information Flow}
\label{subsec:architecture_state}

Beyond recurrence structure and parameter sharing, looped Transformers differ in the information available to the recurrent core at each iteration. Rather than relying solely on the evolving hidden state, Huginn~\citep{geiping2026scaling} provides the recurrent core with a fixed representation of the original input at every iteration. MeSH~\citep{yu2026mesh} further incorporates a memory buffer that can be updated across iterations. These designs distinguish between preserving access to the original input and maintaining additional information that evolves during recurrent computation.

In addition to information flow, the representational capacity of the recurrent state also affects what can be achieved through additional iterations. \citet{saunshi2025reasoning} construct a looped Transformer that reproduces the output of $m$ deterministic chain-of-thought steps in $m$ iterations. However, their construction also uses $m$ additional token positions, a wider hidden representation, and additional attention heads. Thus, its expressivity guarantee cannot be attributed to recurrent depth alone. To assess 	the respective contributions of recurrent depth and workspace size, one can independently vary the number of iterations and workspace positions on tasks with known reasoning depth. Appendix~\ref{app:theory_probes} presents empirical tests for other theoretical results.

\section{Training Recurrent Models}
\label{sec:training}

Training looped Transformers involves several design choices beyond architecture, including parameter initialization, recurrence depth during the forward pass, the hidden states receiving direct supervision, and the iterations through which gradients are propagated. Appendix~\ref{appendix:training} details representative training strategies, while Appendix~\ref{appendix:theory} discusses theoretical learning guarantees and their assumptions on model architecture, data distributions, and optimization procedures.

\subsection{Parameter Initialization and Adaptation}
\label{subsec:training_initialization}

Looped Transformers can be trained from scratch or initialized from pretrained Transformers with untied layers. For example, Huginn~\citep{geiping2026scaling} trains its recurrent core from scratch, whereas Relaxed Recursive Transformers~\citep{bae2025relaxed} initialize a shared core and depth-specific corrections from a pretrained untied model and then undergo further training. When evaluating the training efficiency of models converted from pretrained Transformers, it is important to distinguish the computational cost of adaptation from that of pretraining the original model.

\subsection{Exposure to Recurrence Depth}
\label{subsec:training_depth}

Looped Transformers can be trained with either fixed or variable recurrence depths. For example, Huginn~\citep{geiping2026scaling} samples recurrence depths during training, exposing the recurrent core to different numbers of iterations. The choice of training depths has implications for evaluating recurrent computation at inference. In particular, improvements at larger inference depths do not necessarily demonstrate \emph{recurrence-depth extrapolation}, which requires evaluation at depths beyond those encountered during training. Accordingly, studies investigating recurrence-depth extrapolation should characterize the recurrence depths used during both training and inference.

\begin{figure}[t]
    \centering
    \includegraphics[width=\columnwidth]{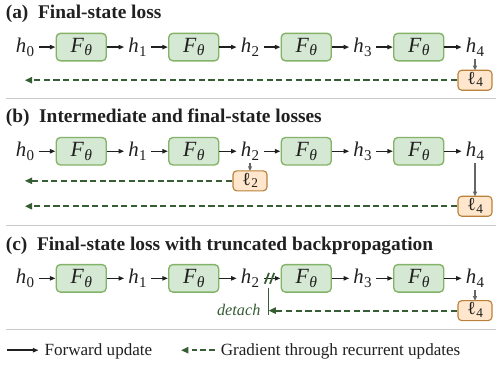}
    \caption{\small Loss placement and gradient propagation across four
    applications of a shared core $F_\theta$. (a) Gradients from the final loss
    propagate through all four updates. (b) An additional loss at $h_2$
    provides gradients through the first two updates. (c) Detaching $h_2$
    restricts backpropagation to the last two updates.}
    \label{fig:training_supervision}
    \vspace{-0.9em}
\end{figure}

\subsection{Supervision and Objectives}
\label{subsec:training_objectives}

Training objectives for looped Transformers differ in where supervision is applied and which supervision signals are used. The placement of supervision depends on the intended use of intermediate states at inference. When only the final prediction is used, supervision can be applied exclusively at the final iteration, allowing intermediate states to evolve without explicit prediction targets, as in Huginn~\citep{geiping2026scaling}. When predictions are required at multiple recurrence depths, intermediate states can be directly supervised using the same next-token prediction objective. For example, Ouro~\citep{zhu2026ouro} weights prediction losses according to learned exit probabilities and regularizes the exit distribution to maintain supervision across depths.

When multiple iterations receive direct supervision, the relative loss weights can affect prediction quality at individual exits. In a two-loop recursive Gemma model, \citet{bae2025relaxed} vary the loss coefficient of the first exit while keeping that of the final exit fixed at 1. Under the same post-training budget of 15 billion tokens, increasing the coefficient from 0.01 to 0.05 improves average few-shot accuracy from 42.1\% to 46.0\% at the first exit and from 52.4\% to 52.6\% at the final exit. Further increasing the coefficient to 0.30 improves first-exit accuracy to 48.5\% but reduces final-exit accuracy to 52.2\%. These results suggest that intermediate supervision can benefit both exits, although placing greater emphasis on early predictions may introduce a trade-off.

Beyond the placement and weighting of supervision, the available training signals determine how intermediate states can be supervised. When annotated reasoning steps are available, RELAY~\citep{yu2026enhancing} trains successive recurrent states to predict the corresponding reasoning steps. Alternatively, RLTT~\citep{williams2026prioritize} uses rewards from completed responses to optimize predictions across recurrent iterations. At a common decoding budget of 2,048 tokens, RLTT improves MATH-500 accuracy from 71.6\% to 86.0\% over final-iteration GRPO on Ouro-2.6B-Thinking. This result illustrates the potential of outcome-based supervision across recurrent iterations to improve final predictions without annotated reasoning steps. Appendix~\ref{appendix:training_losses} provides further details on these objectives and empirical comparisons.

\begin{takeawaybox}
\raggedright
\textbf{Takeaway.}
\emph{Supervision should reflect the intended use of intermediate states at inference and the available training signals. Appropriate loss weighting can improve early predictions without compromising final prediction quality.}
\end{takeawaybox}

\subsection{Gradient Propagation and Optimization}
\label{subsec:training_gradients}

Training looped Transformers with many recurrent iterations can be computationally expensive and introduce optimization challenges. Two important considerations are the depth of gradient propagation and the accumulation of gradients with respect to shared parameters. As illustrated in Figure~\ref{fig:training_supervision}, the placement of prediction losses and the depth of gradient propagation are distinct design choices. Even when supervision is applied only at the final iteration, gradients can propagate through all preceding iterations. Truncating backpropagation reduces computational and memory costs but prevents gradients from propagating through earlier recurrent updates. For example, Huginn~\citep{geiping2026scaling} applies supervision at the final iteration while restricting backpropagation to at most the last eight iterations.

Another challenge arises from parameter sharing: gradient contributions from multiple recurrent iterations accumulate with respect to the same parameters, and a single parameter update affects all applications of the recurrent core. This coupling can affect optimization stability. DeepLoop~\citep{li2026deeploop} addresses this challenge through a parameterization that accounts for gradient accumulation across recurrent iterations.

\section{Inference and Execution}
\label{sec:inference}

Inference in looped Transformers involves three complementary design choices: how recurrence depth is determined, how predictions are obtained from recurrent states, and how computational savings are realized. These choices are distinct: selecting an earlier prediction does not reduce computational cost if subsequent iterations have already been executed. Appendix~\ref{app:inference} provides further technical details on inference and execution strategies.

\subsection{Recurrence Depth Allocation}
\label{subsec:inference_budgets}

The recurrence depth of a looped Transformer can be specified in advance or determined adaptively during inference. Under a fixed-depth policy, the same recurrence budget is applied throughout a given inference run. In some architectures, this budget determines not only the number of iterations but also the computation performed at each iteration. For example, LoopFormer~\citep{jeddi2026loopformer} conditions each recurrent update on the relative progress through the specified recurrence budget and the step size. Thus, changing the recurrence budget can alter the sequence of recurrent updates rather than simply extend or truncate it.

Increasing recurrence depth at inference can improve performance without modifying model parameters. For example, with the same Huginn checkpoint~\citep{geiping2026scaling}, increasing the recurrence count from 8 to 16 and then to 32 iterations improves normalized accuracy on HellaSwag from 58.54\% to 64.67\% and 65.21\%, respectively. However, the smaller gain from 16 to 32 iterations illustrates diminishing returns on this benchmark, motivating a balance between prediction quality and computational cost.

Adaptive policies can allocate recurrent computation at the sequence or token level. At the sequence level, GUT~\citep{chowdhury2025investigating} determines a common stopping point for all tokens in a sequence. At the token level, MoR~\citep{bae2026mixture} investigates two strategies: token-choice routing, which assigns each token a recurrence depth in advance, and expert-choice routing, which selects the $k$ highest-scoring eligible tokens for further computation at each routing step during training. During autoregressive decoding, where future token scores are unavailable, the expert-choice router instead predicts whether the current token would be selected based on the available context.

\begin{figure}[t]
    \centering
    \includegraphics[width=\columnwidth]{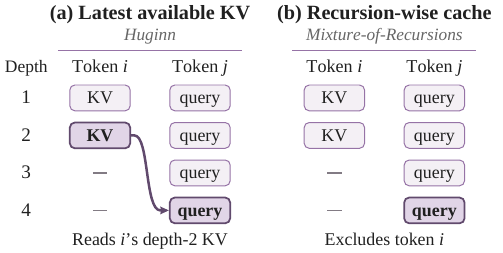}
    \caption{\small Attention after early exit.
    Token $i$ exits at depth 2, while a later token $j$ reaches depth 4.
    (a) Huginn allows $j$ to attend to $i$ using its depth-2 keys and values~\citep{geiping2026scaling}.
    (b) MoR's recursion-wise cache excludes $i$ from $j$'s attention context at depth 4 because no depth-matched KV entry exists~\citep{bae2026mixture}.}
    \label{fig:early_exit_kv_cache}
    \vspace{-0.9em}
\end{figure}

\subsection{Stopping Criteria and Output Selection}
\label{subsec:inference_stopping}

Exit rules determine which recurrent state is used for prediction and can guide the termination of recurrent computation. For example, Huginn~\citep{geiping2026scaling} evaluates a stopping criterion based on changes between consecutive output distributions, whereas Ouro adopts the Q-exit rule introduced by PALBERT, which selects the earliest iteration at which the cumulative exit probability reaches a threshold~\citep{balagansky2022palbert,zhu2026ouro}. However, selecting an earlier prediction does not necessarily reduce executed computation. In the released Ouro implementation examined by \citet{vendrell2026memory}, the threshold determines which logits are returned, but the decoder still executes all remaining iterations.

To distinguish limitations of recurrent computation from those of output selection, \citet{kuo2026stabilizing} evaluate \emph{oracle accuracy} on the Unique Set task, defined as the proportion of examples for which at least one evaluated recurrence depth yields the correct answer. The gap between oracle and policy accuracy indicates the potential improvement through better output selection, whereas low oracle accuracy indicates that correct predictions are rarely available across the evaluated depths.

\begin{figure*}[ht]
\vspace{-0.5em}
    \centering
    \includegraphics[width=0.95\textwidth]{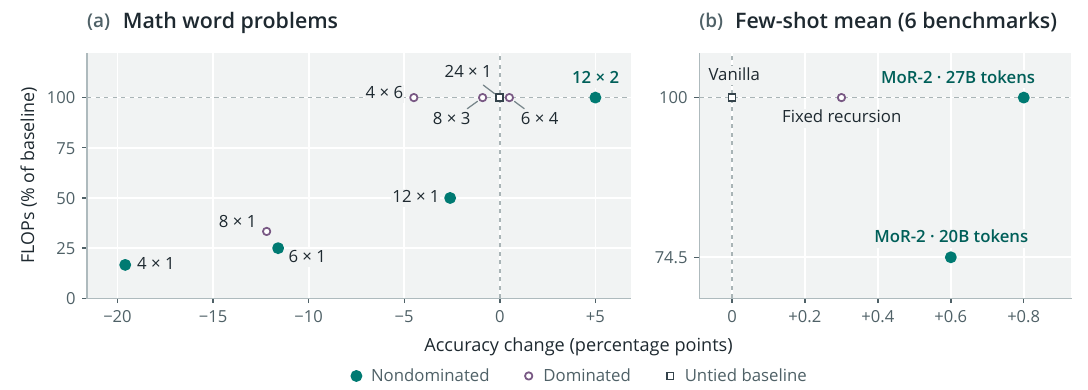}
\caption{\small Accuracy--compute trade-offs relative to untied baselines (squares). Accuracy changes are reported in percentage points and FLOPs as percentages of baseline FLOPs. Filled and open markers denote nondominated and dominated configurations, respectively, within each panel.
\textbf{Left:} All models use 250B training tokens; $k\!\times\!R$ denotes $k$ distinct layers applied $R$ times~\citep{saunshi2025reasoning}.
\textbf{Right:} Vanilla and fixed recursion use 20B training tokens; MoR-2 uses 20B (matched tokens) or 27B (matched forward-pass FLOPs)~\citep{bae2026mixture}.
The panels differ in benchmarks, axis scales, and FLOP accounting.}
    \label{fig:baseline_comparisons}
\vspace{-0.5em}
\end{figure*}

\subsection{Execution and Computational Efficiency}
\label{subsec:inference_execution}

Reducing the number of recurrent updates does not necessarily translate into proportional savings in memory or inference latency. These savings depend on how recurrent computation is implemented, particularly through cache management and execution scheduling. When tokens exit at different recurrence depths, the cache policy determines which representations of earlier tokens remain available to subsequent queries. As illustrated in Figure~\ref{fig:early_exit_kv_cache}, Huginn reuses the latest available keys and values when depth-matched entries are missing, whereas MoR's recursion-wise cache restricts attention to tokens that have reached the query's recurrence depth~\citep{geiping2026scaling,bae2026mixture}. Cache policies therefore affect both memory requirements and the information available to subsequent predictions.

Beyond cache management, variable recurrence depths can reduce hardware utilization by leaving batch positions idle. Continuous Depth Batching~\citep{schwethelm2026depth} addresses this issue by dynamically filling available batch positions with tokens ready to execute the shared block. However, its throughput evaluation on an H100 GPU replays recorded exit decisions and fixes response lengths to those of reference outputs. The gains reported in Table~\ref{tab:survey_synthesis} therefore characterize scheduling efficiency rather than the end-to-end effectiveness of adaptive stopping, which also depends on the overhead of exit decisions and the quality of generated outputs. Appendix~\ref{app:inference} provides further details on caching, scheduling, parallel generation, and verified drafting.

\begin{takeawaybox}
\raggedright
\textbf{Takeaway.}
\emph{Adaptive inference requires joint evaluation of prediction quality, executed computation, and runtime efficiency, as gains in one do not guarantee gains in others.}
\end{takeawaybox}

\begin{table*}[t]
\vspace{-0.9em}
\centering
\small
\setlength{\tabcolsep}{4pt}
\renewcommand{\arraystretch}{1.06}
\begin{tabularx}{\textwidth}{@{}
>{\raggedright\arraybackslash}p{0.19\textwidth}
>{\raggedright\arraybackslash}p{0.24\textwidth}
YY@{}}
\toprule
\textbf{Configuration} & \textbf{Experimental Setting} &
\textbf{Reported Results} & \textbf{Interpretation} \\
\midrule
\multicolumn{4}{@{}l}{\textit{Q1. Effect of inference depth within a fixed checkpoint}} \\

Huginn 3.5B~\citep{geiping2026scaling}
& Same checkpoint evaluated at $R=8,16,32$.
& HellaSwag normalized accuracy: $58.54\%$, $64.67\%$, and $65.21\%$.
& Diminishing accuracy gains at larger depths. Depth extrapolation requires knowledge of the training depths. \\

\addlinespace[0.2em]
\citet{kohli2026loopthink}, four-layer core
& Trained at $R=8$ on tasks with up to 12 reasoning hops; variable inference depth.
& Additional iterations solve some problems requiring more reasoning hops than seen during training.
& Separates generalization to longer reasoning chains from generalization to unseen recurrence depths. \\

\midrule
\multicolumn{4}{@{}l}{\textit{Q2. Model comparisons under controlled resource budgets}} \\

\citet{saunshi2025reasoning}, $12\!\times\!2$ vs.\ 24 untied layers
& Matched training tokens (250B) and executed depth (24 layer applications).
& Average math accuracy: $34.3$ vs.\ $29.3$; validation perplexity: $7.90$ vs.\ $7.40$.
& Parameter sharing improves mathematical reasoning accuracy but worsens language-modeling perplexity in this comparison. \\

\addlinespace[0.2em]
MoR-2, expert-choice~\citep{bae2026mixture}
& MoR-2 vs.\ Vanilla vs.\ fixed recursion, each trained on 20B tokens.
& NLL: $2.7749$, $2.7824$, $2.8079$; FLOPs ($10^{18}$): $12.3$, $16.5$, $16.5$.
& Routed recursion reduces both loss and estimated forward-pass FLOPs, whereas fixed recursion increases loss relative to Vanilla. \\

\addlinespace[0.2em]
DeepLoop, GPT-2 small, $R=7$~\citep{li2026deeploop}
& Same backbone and training tokens (50B) as the pre-normalized baseline.
& Validation cross-entropy: $2.7514$ vs.\ $2.7700$.
& Evaluates the overall parameterization using a single seed; isolating the scaling exponent requires a separate ablation. \\

\midrule
\multicolumn{4}{@{}l}{\textit{Q3. Execution efficiency under measured workloads}} \\

Continuous Depth Batching~\citep{schwethelm2026depth}
& One H100 (80\,GB), recorded exits, and fixed response lengths; full-depth batching baseline.
& Throughput: $1.32$--$1.53\times$ baseline on Ouro 1.4B and $1.30$--$1.37\times$ on Huginn 3.5B at selected thresholds.
& Batch refill is faster on Ouro, whereas no-refill is faster on Huginn at these settings. Timings exclude live exit decisions. \\

\addlinespace[0.2em]
LLoCoT vs.\ Reasoning SFT~\citep{graugarcia2026llocot}
& H100, BF16, batch size 1, 128 coding prompts, three runs.
& Model-execution time: $11.47$ vs.\ $15.16$\,s; peak memory: $6.3$ vs.\ $3.3$\,GiB.
& Lower execution time despite higher peak memory and estimated FLOPs. Output length and quality may differ. \\
\bottomrule
\end{tabularx}
\caption{Representative within-study comparisons of recurrent-depth computation.
Values follow the configuration order listed in each row.
$R$ denotes the recurrence count, and NLL denotes validation negative log-likelihood.}
\label{tab:survey_synthesis}
\vspace{-0.9em}
\end{table*}

\section{From Reported Results to Deployment Decisions}
\label{sec:capabilities}

\subsection{Comparability of Empirical Results}
\label{sec:capabilities_theory}
\label{subsec:empirical_capabilities}

The conclusions that can be drawn from an empirical study depend on what is varied, what is held fixed, and how performance is measured. Table~\ref{tab:survey_synthesis} summarizes seven representative studies according to the questions they address. For example, varying recurrence depth within the same trained model reveals how additional iterations affect prediction quality. In contrast, comparing separately trained models with the same training data and executed depth allows us to examine the effect of parameter sharing. Similarly, scheduling experiments that replay recorded exit decisions measure the efficiency of the scheduling mechanism without accounting for the cost of making these decisions during inference. Appendix~\ref{app:evidence} discusses the experimental conditions and limitations of each study.

To examine accuracy--compute trade-offs while accounting for differences in experimental settings, Figure~\ref{fig:baseline_comparisons} compares configurations from two studies relative to their respective untied baselines. For each configuration, we report the accuracy difference from the baseline ($a-a_0$, in percentage points) and the computational cost relative to the baseline ($100C/C_0$, in percent), where $a_0$ and $C_0$ denote the baseline accuracy and computational cost, respectively. We identify nondominated configurations, for which no other configuration in the same panel offers higher accuracy at the same or lower cost, or lower cost at the same or higher accuracy. Since the two studies differ in evaluation tasks and FLOP accounting, we restrict these comparisons to configurations within each study. All comparisons are based on published point estimates.

\citet{saunshi2025reasoning} examine how parameter sharing affects mathematical reasoning under two comparison settings: a fixed number of distinct layers and a fixed executed depth. All four looped configurations outperform their shallower untied counterparts with the same number of distinct layers. However, the results are mixed when executed depth is held fixed. Compared with an untied 24-layer Transformer, looped configurations that execute 24 layer applications using 12, 8, 6, and 4 distinct layers achieve accuracy differences of $+5.0$, $-0.9$, $+0.5$, and $-4.5$ percentage points, respectively. These results suggest that parameter reuse can improve reasoning accuracy by increasing computational depth without introducing additional layers, but excessive parameter sharing can reduce the benefits of deeper computation.

While the preceding comparison examines parameter sharing at fixed executed depth, adaptive recurrence introduces another possibility: reducing the computation performed per token and reallocating the saved compute during training. In the MoR study~\citep{bae2026mixture}, fixed two-pass recursion improves average few-shot accuracy by 0.3 percentage points over the untied baseline at the same reported FLOP budget, despite increasing language-modeling loss. With adaptive routing, expert-choice MoR-2 achieves a larger accuracy improvement of 0.6 percentage points while using only 74.5\% of the baseline's estimated forward-pass FLOPs when both models are trained on 20B tokens. Under a matched FLOP budget, MoR-2 processes 27B tokens and improves accuracy by 0.8 percentage points. However, these improvements cannot be attributed to parameter sharing alone, since MoR-2 also introduces routing and caching mechanisms and, in the matched-FLOP comparison, processes more training tokens.

The benefits of parameter sharing also vary across downstream tasks. For example, MobileLLM-LS and expert-choice MoR-2 (trained on 20B tokens) improve HellaSwag accuracy over their respective untied baselines by 0.6 and 0.5 percentage points, respectively. On WinoGrande, however, MobileLLM-LS reduces accuracy by 1.0 percentage point, whereas MoR-2 improves it by 2.1 percentage points~\citep{liu2024mobilellm,bae2026mixture}. Direct comparisons across studies remain limited by differences in model scale, training data, depth allocation, and evaluation protocols. In particular, MobileLLM uses zero-shot evaluation, whereas MoR follows benchmark-default prompting and applies byte-length normalization to HellaSwag. These findings motivate evaluating looped Transformers on tasks and under computational constraints representative of their intended deployment.

\subsection{Deployment Decisions}
\label{subsec:design_guide}

The preceding findings suggest that no single looped Transformer configuration is preferable across all tasks and resource constraints. Deployment decisions should therefore begin by identifying the required prediction quality and limiting resource, such as parameter count, runtime memory, or inference latency. Table~\ref{tab:design_diagnostics} summarizes common constraints, relevant empirical evidence, and additional evaluations needed to guide these decisions.

\begin{table*}[t]
\centering
\small
\setlength{\tabcolsep}{4pt}
\renewcommand{\arraystretch}{1.07}
\begin{tabularx}{\textwidth}{@{}>{\raggedright\arraybackslash}p{0.19\textwidth}YY@{}}
\toprule
\textbf{Deployment constraint}
& \textbf{Empirical evidence}
& \textbf{Recommended evaluation} \\
\midrule

Limited parameter storage
& MobileLLM-LS improves average commonsense accuracy by 0.7 percentage points through adjacent layer sharing at the same parameter count (125M)~\citep{liu2024mobilellm}.
& Compare strict sharing with untied models at matched parameter count and executed depth separately. Measure performance on the target device. \\
\addlinespace

Limited peak memory
& On an H100, LLoCoT achieves lower execution time than reasoning SFT (11.47 vs.\ 15.16\,s) but higher peak memory allocation (6.3 vs.\ 3.3\,GiB)~\citep{graugarcia2026llocot}.
& Exclude configurations exceeding the memory budget. Vary prompt length and batch size, accounting separately for weight, cache, and workspace memory. \\
\addlinespace

Adaptation of pretrained models
& Depth-specific low-rank corrections can recover performance lost through parameter sharing, at the cost of additional parameters~\citep{bae2025relaxed}.
& Compare strict and relaxed sharing under the same adaptation budget. Account for additional parameters and evaluate at the intended inference depth. \\
\addlinespace

Variable request difficulty and high concurrency
& Adaptive routing allocates computation selectively across tokens. Replay experiments show that effective batch-refill strategies differ between Ouro and Huginn~\citep{bae2026mixture,schwethelm2026depth}.
& Compare adaptive routing with fixed-depth inference at matched prediction quality. Measure routing and caching overhead, throughput, and 95th-percentile request latency. \\

\bottomrule
\end{tabularx}
\caption{Deployment constraints, supporting empirical evidence, and recommended evaluations. Appendix~\ref{app:design_evaluation} provides additional diagnostics for early-exit accuracy and training cost.}
\label{tab:design_diagnostics}
\end{table*}

\paragraph{Deployment under a parameter budget.}
Parameter sharing can increase computational depth without increasing model size. For example, MobileLLM-125M uses 30 distinct layers, whereas MobileLLM-LS applies each layer twice consecutively. After training on 1T tokens, MobileLLM-LS improves average commonsense accuracy from 46.3\% to 47.0\% at the same parameter count. On an iPhone 13, mean FP16 execution times over 50 runs are 15.6\,ms for MobileLLM, 16.0\,ms for MobileLLM-LS, and 29.0\,ms for a 60-layer untied model~\citep{liu2024mobilellm}. Thus, MobileLLM-LS gains 0.7 percentage points with only 0.4\,ms of additional execution time. However, parameter efficiency does not necessarily imply runtime memory efficiency. If device memory is constrained, peak memory usage, including the attention cache and temporary buffers, should be measured at representative context lengths.

\paragraph{Selecting an inference depth under an accuracy constraint.}
Increasing recurrence depth can improve prediction quality at the cost of inference efficiency. A practical objective is therefore to satisfy the required accuracy while minimizing inference cost. For example, at 8, 16, and 32 recurrent iterations, Huginn achieves normalized HellaSwag accuracies of 58.54\%, 64.67\%, and 65.21\%, respectively. A separate generation benchmark on an RTX 6000 Ada reports corresponding throughputs of 726, 380, and 201 output tokens/s~\citep{geiping2026scaling}. With an accuracy target of 64\%, 16 iterations provide the highest reported throughput among the evaluated depths that satisfy this target. Increasing the recurrence count to 32 improves accuracy by only 0.54 percentage points while nearly halving throughput. To assess further efficiency gains, adaptive policies should be evaluated against this fixed-depth configuration under representative workloads, with measurements of prediction quality, output lengths, throughput, and tail latency.

\paragraph{Choosing between recurrent and alternative computation.}
Although recurrence enables additional computational depth under a fixed parameter budget, alternative approaches may be preferable when this constraint is relaxed or the computational objective differs. If additional parameters can be accommodated, untied Transformers may benefit from distinct layer transformations at the same executed depth, as illustrated in Section~\ref{subsec:empirical_capabilities}. For adaptive computation, conventional early-exit mechanisms~\citep{zhou2020pabee,schuster2022confident} can reduce executed depth without requiring parameter sharing. Unlike recurrent architectures that support depth extrapolation, however, untied Transformers with conventional early-exit mechanisms have a maximum depth bounded by the available layers. Latent reasoning methods offer another alternative by introducing additional workspace through latent token positions~\citep{hao2024training}. Since workspace expansion and recurrent depth can also be combined, as in LLoCoT~\citep{graugarcia2026llocot}, their contributions should be evaluated separately under comparable computational budgets.
\section{Challenges and Future Directions}
\label{sec:future}

Building on the preceding analysis, we identify four challenges in improving the effectiveness and efficiency of looped Transformers across architectural design, training, inference, and deployment.

\paragraph{Architectural design for effective recurrent computation.}
As discussed in Section~\ref{sec:architecture}, increasing computational depth through parameter reuse does not necessarily provide the same benefits as increasing depth through distinct layers. An open question is how to design recurrent cores that effectively utilize additional iterations. Future studies could systematically vary the recurrence structure, parameter-sharing scheme, and state representation under controlled parameter and compute budgets. Such comparisons would help identify which architectural choices enable more effective recurrent computation and under which conditions looped architectures are preferable to untied ones.

\paragraph{Balancing early prediction and iterative refinement.}
Intermediate supervision can improve prediction quality at early exits, potentially making adaptive inference more effective. However, as discussed in Section~\ref{subsec:training_objectives}, improving early predictions may come at the expense of final prediction quality. This raises the question of how to train looped Transformers to produce accurate predictions early while retaining the ability to refine them through additional iterations. Future studies could compare different supervision strategies~\citep{bae2025relaxed,yu2026enhancing} by examining not only accuracy at individual recurrence depths but also the frequency with which subsequent iterations correct existing errors or introduce new ones.

\paragraph{Generalization of adaptive depth allocation.}
Adaptive inference aims to allocate more computation to inputs that benefit from additional recurrent iterations. However, stopping and routing policies may not generalize to inputs of different lengths or reasoning complexities. Future work should evaluate these policies across varying input difficulties without recalibration~\citep{kuo2026stabilizing,bae2026mixture}. Comparing policy accuracy with oracle accuracy can help determine whether poor performance arises from ineffective depth allocation or the model's inability to produce correct predictions across the evaluated depths.

\paragraph{Realizing computational savings in practice.}
The deployment scenarios in Section~\ref{subsec:design_guide} illustrate that reducing recurrent computation does not necessarily translate into practical efficiency gains. These gains also depend on how adaptive depth allocation interacts with cache management and execution scheduling~\citep{bae2026mixture,schwethelm2026depth}. Future work should examine these interactions across different context lengths and levels of request concurrency, comparing peak memory, throughput, and tail latency at matched prediction quality. When adaptive inference requires additional training, it is also important to determine whether the resulting inference savings outweigh the additional training cost (Appendix~\ref{app:design_evaluation}).

\section{Conclusion}
\label{sec:conclusion}

In this paper, we examine looped Transformers along three complementary design axes: architecture, training, and inference. Our analysis highlights that the benefits of recurrent computation depend on the interaction between these design choices, the intended application, and the available computational resources. Parameter sharing can improve model quality under a fixed parameter budget, but its benefits vary across tasks and comparison settings. Appropriate supervision can improve prediction quality across recurrence depths, while adaptive inference can reduce unnecessary computation. However, realizing practical efficiency gains also requires effective cache management and execution scheduling.

By synthesizing theoretical and empirical evidence, we provide practical guidance for designing and evaluating looped Transformers under application-specific performance requirements and resource constraints. Future research should focus on improving the effectiveness of recurrent computation and translating these improvements into reliable predictions and practical efficiency gains.

\clearpage

\section*{Limitations}

We selected representative studies to illustrate model design choices and evaluation practices. Variations in terminology, and our reliance on web-indexed searches and selective reference following, may have limited literature coverage. Appendix~\ref{app:review_procedure} documents the search queries and the available records of the literature identification process.

Our empirical comparisons rely on results reported in the original studies. The plotted frontiers depend on task aggregate, cost metric, and evaluated configurations. Several reported differences are small relative to potential run-to-run variation, and the available evidence does not establish their statistical significance. Moreover, forward-pass FLOPs, peak allocated memory, and end-to-end latency capture different aspects of computational cost. The findings from our deployment case studies are therefore specific to the evaluated hardware and workloads. Generalizing these findings to other settings requires additional measurements of device memory, energy consumption, and tail latency.

The theoretical results discussed in this paper rely on specific assumptions regarding inputs, model size, available workspace, or numerical precision. Their applicability to pretrained language models requires empirically verifying whether the learned recurrent updates exhibit the theoretically characterized behavior.

\section*{Acknowledgments}

This work was supported in part by CoCoSys, one of seven centers in JUMP 2.0, a Semiconductor Research Corporation (SRC) program sponsored by DARPA.

\bibliography{custom}

\clearpage
\appendix

\section{Additional Architectural Details}
\label{appendix:architecture}

This appendix complements Section~\ref{sec:architecture} by formalizing recurrent state transitions, parameter and depth accounting, and selective state updates. We also examine feedback mechanisms and clarify the boundaries of recurrent-depth computation. Table~\ref{tab:architecture_examples} summarizes representative architectures in terms of their recurrent units, parameter-sharing schemes, and state updates or conditioning.

\subsection{Recurrent State and Conditioning}
\label{app:architecture_state_notation}

The recurrence formulation in Section~\ref{sec:definition_scope} combines input processing and state initialization into $\Pbf$. To distinguish these operations and explicitly account for recurrent conditioning, we introduce the following schematic formulation:
\begin{align*}
    \mathbf{c} &= \mathcal{E}(\xbf), \\
    \hbf_0 &= \mathcal{I}(\mathbf{c},\boldsymbol{\epsilon}), \\
    \hbf_{r+1} &=
    \Fbf_{\theta,\phi_r}(\hbf_r;\mathbf{c},\mathbf{s}_r),
    \quad r=0,\ldots,R-1, \\
    \hat{\ybf} &= \Gbf(\hbf_R),
\end{align*}
where $\mathcal{E}$ encodes the input, $\mathcal{I}$ initializes the recurrent state, $\boldsymbol{\epsilon}$ denotes optional initialization noise, and $\mathbf{s}_r$ denotes an iteration-dependent conditioning signal. Unlike the evolving recurrent state $\hbf_r$, the encoded input $\mathbf{c}$ remains fixed and can be accessed at each iteration. For example, $\mathbf{c}$ corresponds to the prompt representation reinjected at each iteration in \citet{yang2024looped} and the fixed prelude representation in Huginn~\citep{geiping2026scaling}, which also employs random state initialization. The conditioning signal $\mathbf{s}_r$ is distinct from $\phi_r$, which denotes learned iteration-specific parameters. 



\paragraph{State organization and sequence resolution.}
The recurrent state need not consist of a single hidden-state tensor. MeSH~\citep{yu2026mesh} uses a multi-slot memory buffer with iteration-specific read and write routers to construct the recurrent core input. Hyperloop Transformers~\citep{zeitoun2026hyperloop} maintain parallel residual streams around a shared recurrent block with loop-specific hyper-connections and positional embeddings. Both architectures enable multiple information propagation pathways while retaining shared computation.

While these architectures modify how recurrent states are organized and propagated, others modify the sequence resolution at which recurrent computation operates. SpiralFormer~\citep{yu2026spiralformer} applies a shared core to compressed latent sequences at different resolutions and maps the resulting updates back to token resolution. Its causal alignment mechanism prevents information from future tokens from leaking into earlier predictions. Consequently, the number of iterations alone does not determine the number of positions processed during recurrent computation.

\subsection{Parameter and Depth Accounting}
\label{app:architecture_accounting}

\paragraph{Depth and reuse patterns.}
Consider a model with $L_{\mathrm{in}}$ non-recurrent input layers, an $L$-layer recurrent core applied $R$ times, and $L_{\mathrm{out}}$ non-recurrent output layers. Under full execution, the nominal computational depth is
\begin{equation}
    D_{\mathrm{nominal}}
    = L_{\mathrm{in}} + RL + L_{\mathrm{out}},
\end{equation}
where $RL$ is the effective recurrent depth defined in Section~\ref{sec:definition_scope}. Adaptive routing or halting may reduce the realized depth for individual inputs or tokens.

This formulation describes a single recurrent core. Architectures with multiple recurrent components may require separate depth accounting, as in machine translation models that apply block-level recurrence independently within the encoder and decoder~\citep{li2021recurrent}. Moreover, identical depth and parameter counts do not imply identical computation. The reuse patterns $A,B,A,B$ and $A,A,B,B$, discussed in Section~\ref{subsec:recurrence_structure}, correspond to \emph{cycle} and \emph{sequence} sharing, respectively~\citep{takase-kiyono-2023-lessons}.

\paragraph{Parameter accounting and depth extrapolation.}
Relaxed parameter sharing introduces iteration-specific parameters alongside shared parameters, as in depth-specific low-rank adaptation~\citep{hu2021lora,bae2025relaxed}. Consider a shared weight matrix $W\in\mathbb{R}^{d_{\mathrm{out}}\times d_{\mathrm{in}}}$ with separate low-rank corrections of rank $k$ at each of the $R$ iterations:
\begin{equation}
    W_r = W + B_rA_r,
\end{equation}
where $B_r\in\mathbb{R}^{d_{\mathrm{out}}\times k}$ and $A_r\in\mathbb{R}^{k\times d_{\mathrm{in}}}$. The total parameter count is
\begin{equation}
    d_{\mathrm{out}}d_{\mathrm{in}}
    + Rk(d_{\mathrm{out}}+d_{\mathrm{in}}),
\end{equation}
excluding biases and other components. Setting all corrections to zero recovers the strictly shared transformation, but eliminating their parameter overhead requires removing the correction matrices. Strict sharing of the entire core additionally requires that all other core parameters remain shared across iterations.

Iteration-specific parameters also constrain depth extrapolation. Under strict sharing, the same core parameters can be reused beyond the training depth, provided that other architectural components support such an extension. In contrast, relaxed sharing with a finite set of learned iteration-specific parameters $\{\phi_r\}$ requires a rule specifying which parameters to use at unseen iterations.

\begin{table*}[t]
\centering
\small
\setlength{\tabcolsep}{4pt}
\renewcommand{\arraystretch}{1.12}

\begin{tabularx}{\textwidth}{@{}
>{\raggedright\arraybackslash}p{0.27\textwidth}
>{\raggedright\arraybackslash}p{0.15\textwidth}
>{\raggedright\arraybackslash}X
>{\raggedright\arraybackslash}X@{}}
\toprule
\textbf{Model} &
\textbf{Repeated Unit} &
\textbf{Parameter Sharing} &
\textbf{State Updates and Conditioning} \\
\midrule

Universal Transformer~\citep{dehghani2018universal}
& Single Transformer layer
& Shared recurrent parameters
& Position and timestep encodings \\
\addlinespace[3pt]

Gated Universal Transformer~\citep{chowdhury2025investigating}
& Single Transformer block
& Shared core with learned gating and halting modules
& Position-wise gated updates; global transition-aware halting \\
\addlinespace[3pt]

ALBERT (fully shared)~\citep{lan2019albert}
& Encoder layer
& Shared attention and feed-forward parameters
& Hidden-state propagation from input embeddings \\
\addlinespace[3pt]

Subformer~\citep{reid2021subformer}
& Central Transformer layer
& Shared central layer; independent boundary layers
& Projections between central and boundary representations \\
\addlinespace[3pt]

RingFormer~\citep{heo2025ringformer}
& Single Transformer layer
& Shared base; level-specific components
& Input-dependent level signals and normalization \\
\addlinespace[3pt]

MoEUT~\citep{csordas2024moeut}
& Multi-layer recurrent group
& Corresponding layers shared across iterations
& Input-dependent expert selection \\
\addlinespace[3pt]

Relaxed Recursive Transformer~\citep{bae2025relaxed}
& Multi-layer block
& Shared base; depth-specific low-rank corrections
& Hidden-state propagation through the core \\
\addlinespace[3pt]

Huginn~\citep{geiping2026scaling}
& Internal recurrent core
& Shared core; independent prelude and coda
& Random state initialization; persistent input conditioning \\
\addlinespace[3pt]

MoR (expert-choice)~\citep{bae2026mixture}
& Token-selective recurrent block
& Shared core; recursion-specific routers
& Routing-weighted updates; preservation of bypassed states \\
\addlinespace[3pt]

DiscoLoop~\citep{fu2026discoloop}
& Transformer backbone
& Shared backbone; shared gate in the learned-gate variant
& Hidden-state propagation with vocabulary-derived feedback \\
\addlinespace[3pt]

Parcae~\citep{prairie2026parcae}
& Internal recurrent core
& Shared core with constrained state-mixing parameters
& Stability-constrained state mixing; normalized input injection \\
\addlinespace[3pt]

HRM-Text~\citep{wang2026hrm}
& Coupled low- and high-level modules
& Parameter sharing within each module across updates
& Separate states; multiple low-level updates per high-level update \\
\addlinespace[3pt]

Encode--Think--Decode~\citep{koishekenov2025etd}
& Internal layer block
& Shared internal block; surrounding layers retained
& Iterative refinement of latent representations \\
\addlinespace[3pt]

ITT~\citep{chen2025inner}
& Token-selective layer
& Shared layer weights; learned thinking-step encodings
& Adaptive token routing and residual accumulation \\
\addlinespace[3pt]

MeSH~\citep{yu2026mesh}
& Multi-layer recurrent core
& Shared core; iteration-specific read/write routers
& Multi-slot memory buffer with routed updates \\
\addlinespace[3pt]

SpiralFormer~\citep{yu2026spiralformer}
& Multi-resolution recurrent core
& Shared core; iteration-specific resampling modules
& Multi-resolution updates with causal alignment \\
\addlinespace[3pt]

Hyperloop Transformer~\citep{zeitoun2026hyperloop}
& Internal recurrent block
& Shared core; loop-specific hyper-connections and embeddings
& Parallel residual streams updated across iterations \\
\addlinespace[3pt]

CoTFormer (fixed-repeat)~\citep{mohtashami2025cotformer}
& Multi-layer Transformer block
& Shared block parameters
& Earlier recurrent representations retained as attention context \\
\addlinespace[3pt]

RINS (basic deterministic)~\citep{alabdulmohsin2025rins}
& Early model block
& Shared early-block parameters
& Recurrent output passed to the later block \\
\bottomrule
\end{tabularx}

\caption{Representative architectural configurations across the three design dimensions in Section~\ref{sec:architecture}. Parameter sharing is characterized at the recurrent core level, with boundary modules and controllers noted where relevant.}
\label{tab:architecture_examples}
\end{table*}

\subsection{Selective State Updates}
\label{app:architecture_updates}

Adaptive routing can reduce the realized depth of individual tokens by selectively updating their states at each iteration. A schematic formulation of selective state updates is given by
\begin{equation}
    \mathbf{h}_{r+1,i}
    =
    \mathbf{h}_{r,i}
    +
    m_{r,i}g_{r,i}
    \mathcal{U}_{\theta}
    (\hbf_r;\mathbf{m}_r)_i,
\end{equation}
where $m_{r,i}\in\{0,1\}$ indicates whether token $i$ is selected for an update, $g_{r,i}$ determines the update weight, and $\mathcal{U}_{\theta}$ denotes the update function. When $m_{r,i}=0$, the corresponding token state remains unchanged. The selection mask $\mathbf{m}_r$ may also determine which tokens participate in attention. Importantly, selective updates reduce computational cost only when the corresponding operations are skipped; masking the results of fully computed updates does not yield such savings.

The expert-choice variant of MoR~\citep{bae2026mixture} implements selective updates by weighting the updates of selected tokens while preserving the states of bypassed tokens. Inner Thinking Transformer (ITT)~\citep{chen2025inner} combines adaptive token routing with residual accumulation and learned thinking-step encodings, allowing selected tokens to undergo additional computation within designated layers. The above formulation captures the general principle of selective updates rather than the exact transition rules of these methods.

\subsection{Feedback Interfaces and the Survey Boundary}
\label{app:architecture_feedback}

Architectures that repeatedly invoke a Transformer do not necessarily perform recurrent-depth computation. As defined in Section~\ref{sec:definition_scope}, inclusion in our survey depends on whether shared Transformer layers repeatedly update an evolving hidden representation along network depth. We examine several feedback mechanisms and related architectures to clarify this distinction.

\paragraph{Continuous vocabulary-derived feedback.}
Looped architectures can incorporate vocabulary-derived feedback without generating discrete tokens. Given a hidden representation $\mathbf{u}$, let $p(v\mid\mathbf{u})$ denote a distribution over the vocabulary $\mathcal{V}$ and $\mathbf{e}_v$ the embedding of token $v$. Such feedback can be expressed as a probability-weighted embedding:
\begin{equation}
    \Phi(\mathbf{u})
    =
    \sum_{v\in\mathcal{V}}
    p(v\mid\mathbf{u})\mathbf{e}_v.
\end{equation}

DiscoLoop~\citep{fu2026discoloop} incorporates a gated and normalized form of this feedback into the hidden state passed between successive backbone applications, preserving the direct hidden-state path. In contrast, PonderLM~\citep{zeng2026ponderlm} accumulates probability-weighted embeddings in the input representation supplied to subsequent model applications. These methods illustrate that continuous vocabulary-derived feedback can operate through different state representations. The use of such feedback alone does not determine whether an architecture satisfies our inclusion criterion.

\paragraph{Related forms of additional computation.}
Additional computation can also be introduced along dimensions other than network depth. Transformer-XL~\citep{dai2019transformerxl} propagates cached representations across input segments, whereas scratchpads and chain-of-thought methods~\citep{nye2021show,wei2022chain} externalize intermediate computation through generated tokens. Neither mechanism inherently constitutes recurrent-depth computation.

\paragraph{Latent workspaces and equilibrium formulations.}
The same inclusion criterion applies to architectures that employ latent workspaces or equilibrium formulations. LLoCoT~\citep{graugarcia2026llocot} repeatedly updates a fixed set of latent positions through a shared Transformer before passing the resulting latent thoughts to an autoregressive answer decoder. Its recurrent workspace falls within our scope, whereas non-autoregressive latent prediction alone does not constitute recurrent-depth computation. FPRM~\citep{movahedi2026fixedpoint} connects explicitly iterated Transformer updates to an equilibrium formulation. We include its explicitly executed recurrent computation but treat purely implicit equilibrium formulations as background.



\section{Additional Details on Training}
\label{appendix:training}

\subsection{Representative Training Strategies}
\label{appendix:training_recipes}

To complement Section~\ref{sec:training}, we summarize representative training strategies in terms of parameter initialization, recurrence depth, supervision objectives, and gradient propagation. Table~\ref{tab:training_lm_recipes} covers language model training and adaptation from pretrained models, while Table~\ref{tab:training_specialized_recipes} presents additional training schedules and supervision targets. Recurrence counts are reported according to the recurrent unit defined in each study. Entries marked as \emph{not specified} indicate that the original study does not clearly specify the gradient truncation policy.

When interpreting recurrence-depth generalization, it is important to distinguish recurrent depth from the depth of an untied network. For example, LayerDrop~\citep{fan2020layerdrop} uses structured layer dropout to enable shallower subnetworks. However, selecting layers from an untied Transformer differs from varying the number of applications of a shared recurrent core. Robustness to reduced network depth therefore does not establish recurrence-depth extrapolation.

\begin{table*}[t]
\vspace{-0.5em}
\centering
\small
\setlength{\tabcolsep}{4pt}
\renewcommand{\arraystretch}{1.12}
\begin{tabularx}{\textwidth}{@{}
>{\raggedright\arraybackslash}p{0.24\textwidth}
>{\raggedright\arraybackslash}X
>{\raggedright\arraybackslash}X
>{\raggedright\arraybackslash}X@{}}
\toprule
\textbf{Study and initialization}
& \textbf{Recurrence depth}
& \textbf{Supervision}
& \textbf{Optimization and other details} \\
\midrule

Huginn~\citep{geiping2026scaling}; from scratch
& Poisson--lognormal depth sampling ($\bar r=32$).
& Final-iteration next-token loss.
& Backpropagation through at most the last eight iterations; activation checkpointing and input reinjection. \\
\addlinespace

Recursive / Relaxed Recursive Transformers~\citep{bae2025relaxed}; conversion
& Typically two iterations; three-iteration variants also explored.
& Language-modeling loss; optional knowledge distillation.
& Joint optimization of shared weights and depth-specific low-rank corrections; separate early-exit training. \\
\addlinespace

Retrofitted recurrence~\citep{mcleish2025teaching}; conversion
& Depth sampling with a scheduled mean.
& Final-iteration next-token loss.
& Backpropagation through at most the last eight iterations; backward-horizon scheduling also explored. \\
\addlinespace

Ouro~\citep{zhu2026ouro}; from scratch
& Eight iterations followed by four; learned exit distribution.
& Exit-weighted next-token losses.
& Entropy regularization followed by gate-only refinement; gradient truncation not specified. \\
\addlinespace

MoR~\citep{bae2026mixture}; from scratch
& Token-dependent depths under depth or capacity constraints.
& Next-token loss.
& Causal routing supervision for expert-choice routing; load balancing or router-bias updates for token-choice routing. \\
\addlinespace

Closing the Loop~\citep{marchenko2026closing}; from scratch and conversion
& Four iterations for training from scratch; three for conversion.
& Exit-weighted losses; smoothed weights for conversion.
& Learning-rate warmup and gate regularization; no step-specific adapters required for conversion. \\
\addlinespace

Training-free retrofit~\citep{chen2026trainingfree}; frozen checkpoint
& Repeated application of selected middle layers at inference.
& No additional training loss.
& Damped recurrent updates; no parameter updates. \\
\addlinespace

Parcae~\citep{prairie2026parcae}; recurrent pretraining
& Variable depths in training and inference studies.
& Language-modeling loss.
& Constrained input-injection parameters; depth-sampling and stability analyses. \\
\addlinespace

Encode-Think-Decode~\citep{koishekenov2025etd}; mid-training
& Fixed recurrence depth per configuration.
& Original language-modeling objective and mid-training data.
& Separate mid-training from a stage-one checkpoint; no additional parameters in the fixed-depth variant. \\

\bottomrule
\end{tabularx}
\caption{Representative training and adaptation strategies for looped language models. Initialization refers to parameter provenance rather than recurrent-state initialization. The training-free approach is included for comparison.}
\label{tab:training_lm_recipes}
\vspace{-0.5em}
\end{table*}

\begin{table*}[t]
\vspace{-0.5em}
\centering
\small
\setlength{\tabcolsep}{4pt}
\renewcommand{\arraystretch}{1.12}
\begin{tabularx}{\textwidth}{@{}
>{\raggedright\arraybackslash}p{0.19\textwidth}
>{\raggedright\arraybackslash}X
>{\raggedright\arraybackslash}X
>{\raggedright\arraybackslash}X@{}}
\toprule
\textbf{Study}
& \textbf{Recurrence depth}
& \textbf{Supervision}
& \textbf{Optimization and other details} \\
\midrule

\citet{yang2024looped}
& Scheduled depth on synthetic in-context learning tasks.
& Shared prediction target across the final window of iterations.
& Loss and backpropagation restricted to the final window. \\
\addlinespace

\citet{fan2024looped}
& Example-specific depth determined by task step counts.
& Final-answer loss at the prescribed depth; no intermediate-state labels.
& Task-derived inference depth; maximum-confidence output selection. \\
\addlinespace

RL-Halting~\citep{kuo2026stabilizing}
& Unrolling to the maximum training depth to construct the stopping distribution.
& Task loss at one sampled exit per example.
& Policy-gradient updates with sampled task loss and entropy regularization. \\
\addlinespace

RELAY~\citep{yu2026enhancing}
& Depth determined by algorithmic problem complexity.
& Iteration-specific reasoning targets and final-answer loss.
& Separate prediction heads for intermediate reasoning steps and final answers. \\
\addlinespace

TRM-Att~\citep{jolicoeur2025less}
& Multiple recursions per supervision segment.
& Answer and halting losses at segment boundaries.
& Backpropagation through only the final complete recursion; states detached between segments. \\
\addlinespace

LoopFormer~\citep{jeddi2026loopformer}
& Full and sampled shorter conditioned trajectories.
& Next-token losses at both trajectory endpoints.
& Consistency loss between trajectory endpoints with stop-gradient. \\
\addlinespace

HRM-Text~\citep{wang2026hrm}
& Two cycles, each with three low-level and one high-level update.
& Response-token loss from the final high-level state.
& Backpropagation horizon increases from two to five module applications. \\
\addlinespace

LoopMTP~\citep{shomali2026loopmtp}
& Fixed depth per configuration.
& Next-token loss on aggregated states; future-token embedding alignment.
& Stop-gradient alignment targets; aggregation gates without early exits. \\
\addlinespace

RLTT~\citep{williams2026prioritize}
& Four-loop Ouro models during post-training.
& Outcome-based advantage applied to weighted per-loop policy terms.
& Final-loop rollouts; no step-level reward annotations required. \\
\addlinespace

LOTUS~\citep{fan2026lotus}
& Recurrent refinement of padded latent positions.
& Parallel CoT-token supervision at post-loop latent positions; answer prediction loss.
& CoT steps correspond to latent positions rather than recurrent iterations. \\

\bottomrule
\end{tabularx}
\caption{Representative methods with different recurrence schedules, supervision objectives, and optimization strategies. Results on synthetic and structured tasks should be interpreted within their experimental settings.}
\label{tab:training_specialized_recipes}
\vspace{-0.5em}
\end{table*}

\vspace{-0.2em}
\subsection{Supervision Objectives and Computation Allocation}
\label{appendix:training_losses}
\vspace{-0.1em}

Training objectives for looped Transformers differ in how supervision is distributed across recurrent iterations, how intermediate states contribute to predictions, and how computation allocation is optimized. We discuss representative formulations to clarify these distinctions.

\paragraph{Related supervision mechanisms.}
Deeply-Supervised Nets introduce auxiliary losses at intermediate layers~\citep{lee2015deeply}, whereas knowledge distillation uses teacher predictions as supervision targets~\citep{hinton2015distilling}. These mechanisms illustrate two complementary aspects of supervision: where it is applied and which targets are used. Although neither mechanism inherently involves recurrent computation, both provide a foundation for understanding supervision across recurrent iterations.

\paragraph{Trade-off between early and final predictions.}
When supervision is applied at multiple depths, the relative loss weights can affect prediction quality at individual exits. In the static-exit ablation of \citet{bae2025relaxed}, a recursive Gemma model with 0.99B non-embedding parameters applies a nine-layer block twice. Following uptraining on 15B tokens, the compared configurations use the same post-training budget of 15B tokens, overlapping training data, and no knowledge distillation. The training objective is a weighted sum of next-token cross-entropy losses computed after the ninth and eighteenth layer applications.

With the final-exit loss coefficient fixed at 1, first-exit coefficients of 0.01, 0.05, 0.10, and 0.30 yield first-exit accuracies of 42.1\%, 46.0\%, 47.1\%, and 48.5\%, respectively, and corresponding final-exit accuracies of 52.4\%, 52.6\%, 52.5\%, and 52.2\%. Accuracy at each exit is averaged across seven few-shot tasks. These results illustrate that increasing the weight of intermediate supervision can improve early predictions while largely preserving final prediction quality within a moderate range. However, the preferred weighting depends on how frequently each exit is used during inference. Fixed-exit evaluations alone do not establish the effectiveness of an adaptive stopping policy.

\paragraph{Exit-weighted supervision.}
Rather than assigning fixed loss weights to individual exits, probabilistic halting methods can weight prediction losses according to learned exit probabilities. For example, PonderNet~\citep{banino2021pondernet} minimizes the expected prediction loss while regularizing the halting distribution toward a geometric prior.

Let $p_{i,r}$ denote the probability of exiting at iteration $r$ for token position $i$, with $\sum_{r=1}^{R}p_{i,r}=1$, and let $\ell_{i,r}$ denote the corresponding next-token loss. Omitting normalization factors, the training objective of Ouro~\citep{zhu2026ouro} is
\begin{equation}
\begin{aligned}
\mathcal{L}_{\mathrm{exit}}
&=\sum_i\left[
\sum_{r=1}^{R}p_{i,r}\ell_{i,r}
-\beta H(\mathbf{p}_i)
\right],\\
H(\mathbf{p}_i)&=-\sum_{r=1}^{R}p_{i,r}\log p_{i,r}.
\end{aligned}
\label{eq:training_exit_entropy}
\end{equation}
The first term minimizes the expected next-token loss under the learned exit distribution, while the entropy regularizer discourages the distribution from concentrating on a single exit, thereby promoting supervision across recurrence depths. The coefficient $\beta$ controls the relative strength of entropy regularization. Gradients propagate through the exit probabilities, allowing token prediction and computation allocation to be optimized jointly. In a subsequent gate-only training stage, detached positive improvements in language-modeling loss are used to construct continuation targets.

For pretrained-model conversion, \citet{marchenko2026closing} smooth the loss weights as
\begin{equation}
\alpha_{r,i}
=(1-\varepsilon)p_{i,r}+\frac{\varepsilon}{R},
\qquad 0<\varepsilon<1,
\label{eq:training_smoothed_exits}
\end{equation}
where $\varepsilon$ controls the contribution of uniform weighting. By assigning each exit a minimum loss weight of $\varepsilon/R$, this formulation ensures nonzero direct supervision at every recurrence depth, regardless of the learned exit probabilities. This smoothing modifies the training objective rather than the inference-time exit rule.

\paragraph{State aggregation.}
Rather than weighting prediction losses at different recurrence depths, state aggregation combines intermediate representations or predictions before computing the prediction loss. For example, ACT~\citep{graves2016adaptive} aggregates intermediate states and introduces a ponder cost to penalize additional computation. LoopMTP~\citep{shomali2026loopmtp} combines recurrent states for next-token prediction while applying auxiliary alignment losses to individual states. These approaches illustrate that aggregation weights can determine how intermediate states contribute to the final prediction without necessarily controlling the number of recurrent iterations performed.


\paragraph{Reward-based learning.}
Beyond prediction losses, reward-based objectives can improve predictions across recurrent iterations or optimize stopping decisions. RLTT~\citep{williams2026prioritize} generates responses from the final-iteration prediction distribution and computes an advantage from each completed response's reward. Unlike standard GRPO, which applies the policy objective only at the final iteration, RLTT uses the advantage to weight policy terms across multiple iterations. This provides direct outcome-based supervision at intermediate iterations without annotated reasoning steps. However, the shared advantage does not explicitly identify the contribution of individual recurrent updates to the final reward.

In the comparison reported by \citet{williams2026prioritize}, RLTT and GRPO are applied to Ouro-2.6B-Thinking under the same training protocol, including four recurrent iterations and 140 optimization steps. At matched maximum decoding lengths of 1,024, 2,048, 3,072, and 4,096 tokens, GRPO achieves MATH-500 accuracies of 42.4\%, 71.6\%, 79.0\%, and 80.8\%, whereas RLTT achieves 78.4\%, 86.0\%, 87.4\%, and 89.8\%, respectively. Although these results do not establish equivalent realized computational costs, they show that distributing the outcome-based learning signal across recurrent iterations can improve reasoning performance across different decoding budgets.

Reward-based learning can also optimize computation allocation rather than prediction quality directly. For example, RL-Halting~\citep{kuo2026stabilizing} trains a stopping policy using the negative task loss at a sampled exit as the reward, with a baseline and entropy regularization. Although task supervision is applied at a sampled iteration, RL-Halting executes all iterations up to the maximum training depth to construct the stopping distribution. Thus, reward-based objectives for improving recurrent predictions and those for learning when to stop address distinct training objectives.

\subsection{Gradient Truncation and Segment-Based Training}
\label{appendix:training_gradient_details}

The repeated application of shared parameters in looped Transformers raises concerns about exploding and vanishing gradients, which have been extensively studied in recurrent networks~\citep{pascanu2013difficulty}. Although these findings do not directly establish the training stability of looped Transformers, they motivate examining how gradients propagate across recurrent iterations. One important consideration is the interaction between gradient truncation and architectural information flow. For example, Huginn~\citep{geiping2026scaling} provides the recurrent core with the input representation produced by the prelude at every iteration. Consequently, gradients can still propagate to the prelude through the retained iterations, even when the recurrent state at the truncation boundary is detached. In contrast, if the input representation is provided only at the initial iteration, detaching an intermediate recurrent state blocks the gradient path to the prelude through that state. Thus, the effect of gradient truncation depends on both the number of retained iterations and the architectural design of information flow.

In addition to controlling gradient propagation, truncation can reduce the memory cost of training. When backpropagation is restricted to a fixed number of final iterations and earlier computation graphs are discarded, the memory required for recurrent activations does not grow with the total number of forward iterations. However, forward computation still increases with recurrence depth. Activation checkpointing offers a complementary approach by recomputing activations during backpropagation~\citep{chen2016checkpointing}. Unlike gradient truncation, checkpointing reduces memory consumption without restricting gradient propagation.

Gradient truncation can also be combined with parameter updates between groups of recurrent iterations. For example, TRM-Att~\citep{jolicoeur2025less} organizes recurrent computation into supervision segments, each containing multiple recurrent updates. At the end of each segment, gradients are propagated only through the final complete recursion, followed by a parameter update. The resulting recurrent states are detached and passed to the next segment. Unlike intermediate supervision within a single unrolled forward pass, this strategy updates model parameters between supervision segments.



\subsection{Parameterization and Stability}
\label{app:training_parameterization}

In looped Transformers, gradient contributions from different recurrent iterations accumulate with respect to shared parameters, whose updates affect all applications of the recurrent core. DeepLoop~\citep{li2026deeploop} analyzes how the alignment of these gradient contributions affects the sensitivity of the model to parameter updates. Under its first-order sensitivity assumptions, this alignment determines the appropriate residual-scaling exponent. For its post-normalized architecture with normalized residual-branch inputs, DeepLoop uses a skip scale of $\alpha=(2N)^{1/2}$ and an initialization gain of $\beta=(8N)^{-1/2}$, where $N$ denotes the unrolled Transformer-block depth. The gain is applied to selected weight matrices at initialization rather than at each recurrent iteration. While the main comparison with a pre-normalized baseline evaluates the overall parameterization, a separate sweep over scaling exponents more directly examines the effect of residual scaling.

Beyond sensitivity to parameter updates, model parameterization can also influence the convergence of hidden states across recurrent iterations. For example, FPRM~\citep{movahedi2026fixedpoint} combines pre-normalization and residual scaling with damped recurrent updates and a residual-based stopping criterion. Its theoretical convergence guarantee, however, relies on a sufficient contraction condition that may not be satisfied in practice.

\section{Additional Details on Inference}
\label{app:inference}

This appendix complements Section~\ref{sec:inference} by examining how inference policies determine recurrence depth, select predictions, and affect practical computational efficiency. We first distinguish selected recurrence depth from executed recurrent computation, then discuss stopping criteria, cache organization, execution scheduling, and verification-based decoding. Appendix~\ref{app:evidence_systems} relates these considerations to the computational costs measured in existing systems evaluations.

\subsection{Selected and Executed Recurrence Depth}
\label{app:inference_accounting}

We first consider \emph{contiguous execution}, in which each token position receives every recurrent update until termination. For $N$ token positions, let $R_i$ denote the number of recurrent-core applications executed for position $i$, with $R_i\leq R_{\max}$, where $R_{\max}$ is the maximum allowed recurrence depth. For an $L$-layer recurrent core, the mean executed recurrence depth and total number of token--layer applications are
\begin{equation}
\bar{R}=\frac{1}{N}\sum_{i=1}^{N}R_i,
\qquad
W_{\mathrm{core}}=L\sum_{i=1}^{N}R_i.
\label{eq:inference_work}
\end{equation}
Neither quantity directly determines FLOPs or latency, which also depend on attention context, boundary modules, intermediate readouts, stopping decisions, and execution overhead.

Let $R_i^{\mathrm{sel}}$ denote the recurrence index of the state selected for prediction. Under contiguous execution, if computation terminates at the selected recurrence, then $R_i^{\mathrm{sel}}=R_i$. However, a policy may evaluate the full trajectory before selecting an earlier state, resulting in $R_i^{\mathrm{sel}}<R_i$ without reducing executed recurrent work. MELT~\citep{vendrell2026memory} documents this distinction in its analysis of an Ouro implementation.

Under \emph{non-contiguous routing}, token positions may skip recurrent updates and resume computation at later iterations. Consequently, recurrence indices no longer directly indicate the number of executed updates. Let $a_{i,r}\in\{0,1\}$ indicate whether the recurrent-core update for position $i$ is executed at iteration $r$. The executed update count for position $i$ and the total number of token--layer applications are then
\begin{equation}
K_i=\sum_{r=1}^{R_{\max}}a_{i,r},
\qquad
W_{\mathrm{core}}=L\sum_{i=1}^{N}K_i.
\label{eq:inference_sparse_work}
\end{equation}
Thus, a state selected at iteration $r$ need not have undergone $r$ recurrent updates. Importantly, $a_{i,r}$ indicates whether an update is executed, regardless of whether the resulting state is retained. These expressions assume that each counted update traverses the full $L$-layer core and exclude padding and execution overhead.

\subsection{Halting Rules and Diagnostic Signals}
\label{app:inference_stopping}

The preceding formulation distinguishes the recurrence selected for prediction from the computation actually executed. We now examine the criteria used to select this recurrence and determine when computation should terminate.

\begin{table*}[t]
\vspace{-0.5em}
\centering
\small
\setlength{\tabcolsep}{4pt}
\renewcommand{\arraystretch}{1.1}
\begin{tabularx}{\textwidth}{@{}
>{\raggedright\arraybackslash}p{0.20\textwidth}
>{\raggedright\arraybackslash}X
>{\raggedright\arraybackslash}X@{}}
\toprule
\textbf{Study} &
\textbf{Early-exit mechanism} &
\textbf{Relationship to recurrent depth} \\
\midrule

DeeBERT~\citep{xin2020deebert}
& Adaptive exits based on intermediate classifiers in BERT.
& Early exiting without recurrent parameter sharing. \\
\addlinespace

FastBERT~\citep{liu2020fastbert}
& Self-distillation of intermediate classifiers for adaptive inference.
& Intermediate supervision without recurrent depth. \\
\addlinespace

PABEE~\citep{zhou2020pabee}
& Patience-based stopping when predictions remain unchanged across consecutive layers.
& Combines early exiting with parameter sharing in its ALBERT configuration. \\
\addlinespace

CALM~\citep{schuster2022confident}
& Token-level exits with sequence-level quality constraints and missing-state handling.
& Stopping and attention mechanisms applicable to recurrent-depth models. \\

\bottomrule
\end{tabularx}
\caption{Representative early-exit methods and their relationship to recurrent-depth Transformers.}
\label{tab:related_early_exits}
\vspace{-0.5em}
\end{table*}

\paragraph{Learned exit distributions.}
While Appendix~\ref{appendix:training_losses} discusses exit probabilities as training-loss weights, we examine their use in selecting recurrence depth at inference. Let $\lambda_{i,r}\in[0,1]$ denote the conditional probability of halting at recurrence $r$ given that position $i$ has reached it. The resulting exit distribution is
\begin{equation}
p_{i,r}
=
\begin{cases}
\lambda_{i,r}
\displaystyle\prod_{s<r}(1-\lambda_{i,s}),
& r<R_{\max}, \\[6pt]
\displaystyle\prod_{s<R_{\max}}(1-\lambda_{i,s}),
& r=R_{\max}.
\end{cases}
\label{eq:inference_exit_distribution}
\end{equation}
The final recurrence receives the remaining probability mass, ensuring $\sum_{r=1}^{R_{\max}}p_{i,r}=1$.

Rather than sampling from this distribution, Ouro adopts Q-exit, a deterministic criterion introduced by PALBERT~\citep{balagansky2022palbert,zhu2026ouro}. Q-exit selects the earliest recurrence at which the cumulative exit probability reaches a threshold $\eta$:
\begin{equation}
R_i^{\mathrm{sel}}
=
\min\left\{
r:
\sum_{s=1}^{r}p_{i,s}\geq\eta
\right\},
\qquad 0<\eta\leq1.
\label{eq:inference_quantile_exit}
\end{equation}
Unlike ACT, which combines halting with state aggregation~\citep{dehghani2018universal}, Q-exit selects a single recurrence. Computational savings require skipping subsequent updates rather than selecting an output after executing them.

\paragraph{Confidence and stability criteria.}
Instead of relying on learned exit probabilities, stopping policies can use signals derived from intermediate predictions or hidden states. Let $\boldsymbol{\pi}_{r,i}$ denote the output distribution at position $i$ after recurrence $r$. Given its two largest probabilities $\pi_{r,i}^{(1)}\geq\pi_{r,i}^{(2)}$, prediction confidence can be characterized by the margin
\begin{equation}
c_{r,i}
=
\pi_{r,i}^{(1)}-\pi_{r,i}^{(2)}.
\label{eq:inference_confidence}
\end{equation}
Prediction stability can be measured by the KL divergence between consecutive output distributions, as explored by Huginn~\citep{geiping2026scaling}:
\begin{equation}
d^{\mathrm{pred}}_{r,i}
=
D_{\mathrm{KL}}\left(
\boldsymbol{\pi}_{r-1,i}
\,\Vert\,
\boldsymbol{\pi}_{r,i}
\right).
\label{eq:inference_prediction_stability}
\end{equation}
Alternatively, Ouro~\citep{zhu2026ouro} evaluates representation stability instead of using its learned gate, measuring the distance between consecutive hidden states:
\begin{equation}
d^{\mathrm{state}}_{r,i}
=
\left\|
\mathbf{h}_{r,i}-\mathbf{h}_{r-1,i}
\right\|_2.
\label{eq:inference_state_stability}
\end{equation}

\paragraph{Second-order latent changes.}
While representation stability measures the magnitude of successive state updates, stopping criteria can also examine how these updates change across iterations. Let $\Delta\hbf_r=\hbf_r-\hbf_{r-1}$ denote the recurrent state update. The criterion of \citet{pappone2025twoscale} measures
\begin{equation}
    a_r=\|\Delta\hbf_r-\Delta\hbf_{r-1}\|_2.
\end{equation}
Their two-hit rule requires the threshold condition to hold at two consecutive updates. Unlike first-order state differences, this measure characterizes changes between successive updates rather than their magnitudes. A small second-order difference therefore does not necessarily imply that the recurrent state has stabilized.

State-based stopping criteria do not require intermediate vocabulary readouts to evaluate the stopping signal but still incur state-comparison overhead. Since neither confidence nor stability guarantees prediction correctness, evaluations should independently assess task performance and report the threshold, evaluation frequency, maximum recurrence depth, and computational overhead.

\paragraph{Relation to early-exit language models.}
Early exiting does not require recurrent parameter sharing. DeeBERT and FastBERT investigate adaptive exits in BERT-based models~\citep{xin2020deebert,liu2020fastbert}, while PABEE employs patience-based stopping with BERT and ALBERT~\citep{zhou2020pabee}. For autoregressive generation, CALM investigates token-dependent early exits, relates local stopping decisions to sequence-level quality constraints, and addresses missing representations of previously exited tokens~\citep{schuster2022confident}. The latter issue also arises in looped Transformers, motivating the cache policies discussed in the following subsection. Table~\ref{tab:related_early_exits} summarizes these methods and their relationship to recurrent-depth.

\subsection{Cache Organization and Attention Semantics}
\label{app:inference_cache}

When tokens undergo different numbers of recurrent updates, their available representations may also differ across recurrence depths. A cache policy determines which keys and values (KVs) are retained and which are available to subsequent queries. Even with shared projection weights, hidden states at different recurrences can produce distinct KVs~\citep{vendrell2026memory,neill2026looped}. Looped Transformers therefore differ in whether they retain, reuse, or reconstruct recurrence-specific KVs.

\paragraph{Cache memory accounting.}
Let $n$ denote the cached sequence length and $L_{\mathrm{att}}$ the number of attention layers in the recurrent core. Let $d_{\mathrm{KV}}=n_{\mathrm{KV}}d_h$ denote the total key width, where $n_{\mathrm{KV}}$ is the number of KV heads and $d_h$ is the head dimension, and $b$ the number of bytes per scalar. Assuming equal key and value widths and uniform KV-head configurations across core layers, storing recurrence-specific KVs for all tokens at fixed depth $R$ requires
\begin{equation}
    M_{\mathrm{core}}^{\mathrm{dense}}
    =
    2bnL_{\mathrm{att}}Rd_{\mathrm{KV}},
    \label{eq:inference_dense_cache}
\end{equation}
excluding boundary layers, metadata, and temporary buffers.

More generally, let $C_i$ denote the number of recurrence-specific KV sets retained for position $i$. Assuming each set contains KVs for all $L_{\mathrm{att}}$ attention layers, the storage requirement is
\begin{equation}
    M_{\mathrm{core}}^{\mathrm{sparse}}
    =
    2bL_{\mathrm{att}}d_{\mathrm{KV}}
    \sum_{i=1}^{n}C_i.
    \label{eq:inference_sparse_cache}
\end{equation}
Because KV entries can be overwritten or reused, $C_i$ need not equal the executed update count $K_i$ in Equation~\eqref{eq:inference_sparse_work}. This formulation does not directly apply to compressed or aggregated representations.

\paragraph{Cache organization and reuse.}
Ouro~\citep{zhu2026ouro} evaluates a last-step KV-sharing policy that reuses final-recurrence KVs of previously generated tokens across recurrent iterations during decoding while retaining recurrence-specific KVs during prefill. In contrast, the recursive KV-sharing variant of MoR~\citep{bae2026mixture} reuses first-recurrence KVs at subsequent depths. MELT~\citep{vendrell2026memory} maintains a gated shared cache alongside recurrence-specific KVs for a bounded window of recent tokens. Looped Latent Attention (LLA)~\citep{neill2026looped} stores compressed representations from which recurrence-specific KVs are reconstructed. These strategies reduce cache storage through reuse, selective retention, or compression.

\paragraph{Attention semantics.}
Beyond memory requirements, cache policies determine how queries access representations when depth-matched KVs are unavailable. Halting a token's updates does not necessarily remove its representation from the attention context. Universal Transformers retain halted representations while other positions continue updating~\citep{dehghani2018universal}. During autoregressive decoding, Huginn~\citep{geiping2026scaling} reuses available KVs from earlier recurrences when depth-matched entries are missing. In contrast, the recursion-wise caching variant of MoR~\citep{bae2026mixture} restricts attention to tokens that have reached the corresponding recurrence depth. Table~\ref{tab:inference_cache} summarizes representative strategies and their implications for attention.

\begin{table*}[t]
\vspace{-0.5em}
\centering
\small
\setlength{\tabcolsep}{4pt}
\renewcommand{\arraystretch}{1.12}
\begin{tabularx}{\textwidth}{
@{}
>{\raggedright\arraybackslash}p{0.25\textwidth}
>{\raggedright\arraybackslash}p{0.35\textwidth}
>{\raggedright\arraybackslash}X
@{}}
\toprule
\textbf{Cache strategy}
&
\textbf{Stored information}
&
\textbf{Attention semantics}
\\
\midrule

Depth-specific caching~\citep{geiping2026scaling}
&
Separate KVs for each retained recurrence.
&
Uses depth-matched KVs when available.
\\
\addlinespace

Cross-recurrence reuse (Ouro)~\citep{zhu2026ouro}
&
Final-recurrence KVs for generated tokens; recurrence-specific KVs for prompt tokens.
&
Reuses final-recurrence KVs of generated tokens across recurrences.
\\
\addlinespace

Shared or aggregated caching~\citep{vendrell2026memory}
&
Shared KV states combining information across recurrences.
&
Attends to aggregated rather than depth-specific representations.
\\
\addlinespace

Compressed caching~\citep{neill2026looped}
&
Compressed latent representations encoding KVs across recurrences.
&
Reconstructs depth-specific KVs, potentially introducing approximation error.
\\

\bottomrule
\end{tabularx}
\caption{Representative KV-cache strategies for looped Transformers, distinguished by their storage mechanisms and attention semantics. These strategies are not mutually exclusive.}
\label{tab:inference_cache}
\vspace{-0.5em}
\end{table*}

\paragraph{Complementary memory optimization techniques.}
Multi-query and grouped-query attention share KVs across query heads rather than recurrent depths~\citep{shazeer2019multiquery,ainslie2023gqa}, reducing $d_{\mathrm{KV}}$ without eliminating recurrence-specific KV states. PagedAttention~\citep{kwon2023efficient} manages KV storage through non-contiguous memory blocks and supports memory sharing across requests. These techniques address complementary aspects of cache efficiency.

\subsection{Scheduling and Parallel Execution}
\label{app:inference_systems}

The preceding measures characterize executed recurrent work and cache storage but do not directly capture how efficiently computation is executed on hardware. Variable recurrence depths introduce scheduling challenges because tokens may require different numbers of updates.

\paragraph{Depth-wise batching.}
Depth-wise batching groups computations at different recurrence depths for execution by the shared core, provided that their recurrent states and required attention inputs are available. Recursive Transformers propose this strategy~\citep{bae2025relaxed}, MoR implements queue-based scheduling~\citep{bae2026mixture}, and Continuous Depth Batching coordinates both recurrent and non-recurrent stages~\citep{schwethelm2026depth}. Non-recurrent modules and iteration-specific parameters may impose additional constraints on which computations can be grouped. A related approach is Orca~\citep{yu2022orca}, which schedules computation at the granularity of autoregressive decoding steps rather than recurrent-depth updates.

\paragraph{Architectural modifications for parallel execution.}
Unlike depth-wise batching, which preserves existing recurrent dependencies, some architectures modify these dependencies to enable parallel execution. Parallel Loop Transformer~\citep{wu2025parallel} allows computation across loop stages to overlap, while LoopCoder-v2~\citep{yang2026loopcoderv2} further explores this architectural design, including shifted inputs between loop stages.

\paragraph{Parallel generation and refinement.}
Beyond parallel execution across recurrent stages, some approaches overlap token generation with latent-state refinement. The diffusion-forcing sampler of \citet{geiping2025parallel} advances left-to-right token generation while continuing to update earlier latent states. Persistent input injection enables these states to incorporate changes in conditioning. This approach reduces sequential dependencies without necessarily reducing FLOPs or reproducing the output distribution of the sequential sampler, introducing a speed--quality trade-off. In contrast, LLoCoT separates joint latent-workspace refinement from autoregressive answer decoding rather than overlapping the two (Appendix~\ref{app:architecture_feedback}).

\subsection{Verification-Based Decoding}
\label{app:inference_verification}

While the preceding approaches improve execution through scheduling or parallelism, verification-based decoding exploits predictions from intermediate recurrent states to accelerate generation. Unlike early-exit decoding, which directly uses shallower predictions as outputs, verification-based decoding treats them as proposals to be accepted or corrected by deeper computation. Speculative decoding preserves the target sampling distribution through its verification and correction procedure~\citep{leviathan2023speculative}, and LoopSpec~\citep{cho2026loopspec} applies this principle to proposals from shallower recurrent states. The resulting speedup depends on the proposal acceptance rate and the computational costs of drafting and verification.
\section{Detailed Theoretical Results and Assumptions}
\label{appendix:theory}

We interpret the theoretical results in Table~\ref{tab:theoretical_results} along two dimensions: how model parameters are obtained and how computational resources scale with input length. Function approximation theorems and algorithmic constructions establish the existence of suitable parameters, whereas learning guarantees specify the conditions under which such parameters can be obtained through optimization. The results also differ in whether computational resources, such as recurrence depth, model width, numerical precision, and workspace, remain fixed or grow with input length. This distinction separates fixed-length function approximation~\citep{yun2020universal} from asymptotic studies of computational expressivity~\citep{hahn2020limitations}. Based on these distinctions, we discuss the implications of theoretical results for architectural design and training (Sections~\ref{sec:architecture} and~\ref{sec:training}) and propose empirical evaluations of trained models.

\subsection{Function Approximation}

\citet{xu2024expressive} study the approximation capabilities of looped Transformers for the class $\mathcal{F}_{\mathrm{PE}}([0,1]^{d\times N})$ of continuous permutation-equivariant functions
\[
f:[0,1]^{d\times N}\rightarrow\mathbb{R}^{d\times N},
\]
at a fixed sequence length $N$, with approximation error measured in the $L^p$ norm for $1\leq p<\infty$. In addition to the standard modulus of continuity $\omega_f$, their analysis introduces the \emph{token modulus} $\omega_f^{\mathrm{tok}}$ and the \emph{contextual modulus} $\omega_f^{\mathrm{cont}}$. These characterize how an output token varies when its corresponding input token or surrounding context changes, respectively, while the other remains fixed.

For any $f\in\mathcal{F}_{\mathrm{PE}}([0,1]^{d\times N})$ and recurrence count $r>N$, the authors construct a recurrent Transformer
\[
\mathrm{TF}: \mathbb{R}^{(17d+9)\times N} \rightarrow \mathbb{R}^{(17d+9)\times N},
\]
with two attention heads of dimension $1$ each and a feed-forward width of $49d+25$, together with affine input and output maps $\mathcal{L}_1$ and $\mathcal{L}_2$, respectively. Let
\begin{equation}
\delta = \left(\frac{r-N}{2}\right)^{-1/((N+1)d+1)}.
\label{eq:xu_delta}
\end{equation}
Under arbitrary numerical precision, the construction satisfies
\begin{align}
&\left\|\mathcal{L}_2 \circ \mathrm{TF}^{\circ r} \circ \mathcal{L}_1 - f\right\|_{L^p}
\nonumber\\
&\quad\leq (Nd)^{1/p}
\left[
\omega_f^{\mathrm{tok}}(\delta\sqrt{d})
+
\omega_f^{\mathrm{cont}}(\delta\sqrt{Nd})
\right] \nonumber\\
&\quad + \omega_f(\delta\sqrt{Nd}) + O\!\left(N^{2/p}\delta^{d/p}\right),
\label{eq:xu_approximation}
\end{align}
where $\mathrm{TF}^{\circ r}$ denotes the $r$-fold composition of the shared Transformer block. When the absolute magnitude of each parameter is bounded by $M$, the approximation bound includes an additional error term:
\begin{equation}
O\!\left(\left((M\delta)^{-1}dN\right)^{1/p}\right).
\end{equation}
The construction has a bit complexity of $O\!\left(\delta^{-(N+1)d}\right)$.

These results demonstrate that increasing recurrence depth can improve approximation accuracy without increasing the architectural size of the recurrent core. Specifically, increasing $r$ decreases $\delta$ and tightens the arbitrary-precision bound, but increases the bit complexity of the construction and need not reduce the additional error term when $M$ is fixed. Moreover, the constructed parameters may depend on the recurrence count and target accuracy. Thus, the universality result characterizes the expressivity of looped Transformers at a fixed sequence length rather than guaranteeing improved accuracy when a fixed trained model executes additional iterations.

\subsection{Constructive Results for Iterative Computation}

Beyond approximating continuous functions, looped Transformers can implement specific iterative algorithms through explicitly constructed parameters. The following results illustrate how different algorithms use recurrent depth, architectural width, numerical precision, and workspace.

\paragraph{Gradient descent in the forward pass.}
\citet{vonoswald2023gradient} show that linear self-attention can implement a gradient-descent update for regression. Building on this result, \citet{chen2025bypassing} construct a linear looped Transformer that performs multiple gradient-descent steps for in-context learning on linear vector-generation tasks. Assuming realizable labels $y=X\theta^*$ with $\|\theta^*\|_2=1$, a zero initial query, nonsingular $X^\top X$, and the prescribed weights and step size $\eta=1/\lambda_{\max}(X^\top X)$, their theorem bounds the error $|\langle v,\theta^*\rangle-\alpha|$ for the generated vector $v$ and query scalar $\alpha\neq0$. The dependence of this bound on the Gram-matrix condition number characterizes the convergence of the forward computation rather than the optimization of model parameters.

\paragraph{Programmable computation.}
\citet{giannou2023looped} construct looped Transformers with fixed, programmed weights that execute instructions encoded in the input sequence. By representing both instructions and memory in the input, their construction illustrates how shared Transformer layers can implement different computational operations through recurrent state updates.

\paragraph{Finite-group composition.}
\citet{saunshi2025reasoning} construct a one-layer looped Transformer that computes the composition of $n$ elements $g_1,\ldots,g_n$ from any finite group $G$,
\[
g_1\circ g_2\circ\cdots\circ g_n,
\]
using $T=\lceil\log_2 n\rceil$ loops. The construction uses constant numerical precision and two attention heads, with dimensions
\begin{align*}
&d = 3\left(
\lceil\log_2|G|\rceil
+
\lceil\log_2 n+1\rceil
\right),\\
&d_{\mathrm{FF}} =
|G|^2+6\lceil\log_2|G|\rceil,\\
&d_{\mathrm{ATTN}} =
\lceil\log_2 n\rceil,
\end{align*}
where $d$, $d_{\mathrm{FF}}$, and $d_{\mathrm{ATTN}}$ denote the embedding, feed-forward, and attention dimensions, respectively. Despite using a single shared layer, the construction requires embedding and attention dimensions that grow logarithmically with input length.

\paragraph{$p$-hop induction.}
\citet{saunshi2025reasoning} also construct a one-layer looped Transformer that solves $p$-hop induction using $\lfloor\log_2 p\rfloor+2$ loops. Unlike the group-composition construction, it maintains constant architectural dimensions,
\[
d=d_{\mathrm{FF}}=d_{\mathrm{ATTN}}=O(1),
\]
with at most three attention heads, but requires $O(\log n)$ bits of numerical precision for input length $n$.

\paragraph{Simulation of chain-of-thought.}
For an $L$-layer untied Transformer with fixed input length $n$, \citet{saunshi2025reasoning} construct a looped Transformer that reproduces the output after $m$ chain-of-thought (CoT) steps generated by deterministic argmax decoding. The construction uses $m$ loops, $m$ additional dummy tokens, a recurrent block with $L+O(1)$ layers, increased embedding dimension, and a constant number of additional attention heads. These resource requirements describe the proposed construction rather than establish lower bounds for CoT simulation.

\subsection{Fully Uniform Computation}

The preceding constructions may require architectural dimensions or parameter values to vary with input length. In contrast, \citet{merrill2026little} study \emph{fully uniform} Transformers, whose architecture and parameters remain fixed while the recurrence count may depend on input length $n$. Their constructions use averaging hard attention and require $\Omega(\log n)$ bits of numerical precision to manipulate token positions. Thus, full uniformity allows computational resources to grow without changing the model architecture or parameters.

\paragraph{Regular-language recognition.}
For any regular language recognized by a finite automaton with state set $Q$, the authors construct a causally masked $(0,8,9)$-universal Transformer, consisting of an eight-layer recurrent block followed by nine non-recurrent layers. Repeating the recurrent block $D(n)=\lceil\log_2 n\rceil$ times yields an effective depth of $8\lceil\log_2 n\rceil+9$. The model dimension and feed-forward width may depend on $|Q|$ but remain independent of input length $n$. Under standard complexity-theoretic conjectures such as $\mathsf{TC}^0\neq\mathsf{NC}^1$, certain regular languages cannot be recognized by fixed-depth Transformers under the corresponding formal model. This separation is conditional, whereas the constructive logarithmic-depth upper bound holds unconditionally.

\paragraph{Graph connectivity.}
For $s$--$t$ connectivity in a graph with $n$ vertices, the authors construct a $(17,2,1)$-universal Transformer with fixed model dimension and feed-forward width. Its two-layer recurrent block is repeated $\lceil\log_2 n\rceil$ times. The construction uses both causal and unmasked attention heads and processes the graph's $n\times n$ adjacency matrix together with $n^3$ padding tokens that serve as computational workspace.

Let $B_\ell(i,j)$ indicate whether $j$ is reachable from $i$ by a path of length at most $2^\ell$. The recurrent computation updates reachability through
\begin{equation}
B_{\ell+1}(i,j)
\iff
\exists k\,
\left[
B_\ell(i,k)\land B_\ell(k,j)
\right].
\end{equation}
Each iteration doubles the path-length bound, allowing connectivity to be determined after $\lceil\log_2 n\rceil$ iterations. Unlike the regular-language construction, however, this result requires polynomially growing computational workspace.

These constructions illustrate how fully uniform Transformers accommodate increasing input lengths through additional recurrent computation and numerical precision. The graph connectivity construction further demonstrates the role of additional workspace, motivating a more general characterization of the computational capabilities afforded by recurrent depth and padding.

\subsection{Computational Complexity with Padding}

The preceding graph connectivity construction uses padding tokens as computational workspace. \citet{merrill2026exact} characterize how recurrent depth and padding jointly determine the class of recognizable languages. Their formal Transformer model uses averaging hard attention, masked pre-normalization, causal masking, a beginning-of-sequence token, and no positional embeddings. The characterization holds under the logarithmic- or polynomial-precision arithmetic models specified in their work.

For integers $d,k\geq0$, let $\mathsf{AHAT}_k^d$ denote the class of languages recognizable by this model with $O(\log^d n)$ recurrent depth and $O(n^k)$ padding tokens. Allowing an arbitrary polynomial amount of padding gives
\begin{equation}
\mathsf{AHAT}_*^d = \bigcup_{k=0}^{\infty}\mathsf{AHAT}_k^d.
\end{equation}

\paragraph{Fixed depth.}
With constant recurrent depth, polynomial padding does not extend computational expressivity beyond constant-depth threshold circuits:
\begin{equation}
\mathsf{AHAT}_*^0 = \mathsf{FO}\text{-uniform }\mathsf{TC}^0.
\label{eq:ahat_tc0}
\end{equation}

\paragraph{Polylogarithmic recurrent depth.}
For every fixed integer $d\geq1$, the authors establish
\begin{equation}
\mathsf{AHAT}_*^d = \mathsf{FO}\text{-uniform }\mathsf{TC}^d,
\label{eq:ahat_tcd}
\end{equation}
where $\mathsf{TC}^d$ denotes polynomial-size threshold-circuit families with depth $O(\log^d n)$. This equality establishes both containment in the corresponding circuit class and the realizability of every language in that class.

Taking the union over all constant values of $d$ yields
\begin{align}
\bigcup_{d=0}^{\infty}\mathsf{AHAT}_*^d
&= \bigcup_{d=0}^{\infty}\mathsf{FO}\text{-uniform }\mathsf{TC}^d
\nonumber\\
&= \mathsf{FO}\text{-uniform }\mathsf{NC}.
\end{align}

These results clarify the complementary roles of padding and recurrence: padding expands the available workspace, whereas recurrence increases sequential computational depth. In particular, polynomial workspace alone is insufficient to move beyond $\mathsf{TC}^0$ at fixed depth, highlighting the importance of growing recurrent depth in the computational characterization.

\begin{table*}[t]
\vspace{-0.5em}
\centering
\small
\setlength{\tabcolsep}{2.5pt}
\renewcommand{\arraystretch}{1.03}
\begin{tabularx}{\textwidth}{
    @{}
    >{\raggedright\arraybackslash}p{0.12\textwidth}
    >{\raggedright\arraybackslash}p{0.14\textwidth}
    >{\raggedright\arraybackslash}p{0.28\textwidth}
    >{\raggedright\arraybackslash}p{0.12\textwidth}
    >{\raggedright\arraybackslash}X
    @{}
}
\toprule
\textbf{Work} &
\textbf{Setting} &
\textbf{Formal guarantee} &
\textbf{Recurrence / depth} &
\textbf{Key assumptions and resources} \\
\midrule

\citet{xu2024expressive}
&
Function approximation
&
Recurrence-dependent approximation bounds and universal approximation of continuous permutation-equivariant functions under the $L^p$ norm ($1\leq p<\infty$).
&
$r>N$; $r$ may grow
&
Fixed sequence length $N$; two attention heads of size $1$; model dimension $17d+9$ and feed-forward width $49d+25$. Universality assumes unrestricted real-valued parameters; bounded parameter magnitudes introduce an additional error term.
\\

\midrule

\citet{saunshi2025reasoning}
&
Finite-group composition
&
Computes the composition of $n$ elements from any finite group $G$ using a single shared layer.
&
$\lceil\log_2 n\rceil$ loops
&
Constant numerical precision; embedding and attention dimensions grow logarithmically with $n$.
\\

\addlinespace

&
$p$-hop induction
&
Solves $p$-hop induction using a single shared layer.
&
$\lfloor\log_2 p\rfloor+2$ loops
&
Constant embedding, feed-forward, and attention dimensions; at most three attention heads; $O(\log n)$ bits of precision.
\\

\addlinespace

&
CoT simulation
&
Reproduces the output after $m$ deterministic argmax CoT steps of an untied Transformer.
&
$m$ loops
&
Fixed input length; $m$ additional workspace tokens; increased embedding dimension and a constant number of additional attention heads.
\\

\midrule

\citet{chen2025bypassing}
&
Linear vector generation
&
Implements $T$ gradient-descent steps in the forward pass, yielding $v$ with $|\langle v,\theta^*\rangle-\alpha|\leq|\alpha|e^{-T/(2\kappa)}$.
&
$T$ loops
&
Realizable labels $y=X\theta^*$, $\|\theta^*\|_2=1$, zero initial query, and $\alpha\neq0$; prescribed weights, mask, and step size. $\kappa$ is the condition number of nonsingular $X^\top X$.
\\

\midrule

\citet{merrill2026little}
&
Regular-language recognition
&
A fully uniform Transformer recognizes any given regular language.
&
$\Theta(\log n)$ depth
&
Fixed architecture and parameters across input lengths; width may depend on the target language; $\Omega(\log n)$ bits of precision.
\\

\addlinespace

&
Graph connectivity
&
A fully uniform Transformer solves directed or undirected $s$--$t$ connectivity.
&
$\lceil\log_2 n\rceil$ loops
&
Fixed model dimension and feed-forward width; $n^3$ padding tokens as workspace; causal and unmasked attention heads.
\\

\midrule

\citet{merrill2026exact}
&
Fixed-depth padded computation
&
Exact characterization of $\mathsf{FO}$-uniform $\mathsf{TC}^0$.
&
$O(1)$ depth
&
Polynomial padding; averaging hard attention, masked pre-normalization, causal masking, a beginning-of-sequence token, and no positional embeddings; specified logarithmic- or polynomial-precision arithmetic.
\\

\addlinespace

&
Polylogarithmic-depth padded computation
&
Exact characterization of $\mathsf{FO}$-uniform $\mathsf{TC}^d$ for fixed $d\geq1$, yielding $\mathsf{FO}$-uniform $\mathsf{NC}$ when taking the union over $d$.
&
$O(\log^d n)$ depth
&
Same formal Transformer model with polynomial padding as workspace.
\\

\bottomrule
\end{tabularx}
\caption{Representative theoretical expressivity guarantees for looped Transformers, with their recurrence requirements and computational assumptions. Learning guarantees are discussed separately in Appendix~\ref{app:theory_learning}.}
\label{tab:theoretical_results}
\vspace{-0.5em}
\end{table*}

\subsection{Learning Guarantees in Restricted Models}
\label{app:theory_learning}

While the preceding results establish that looped Transformers can implement iterative algorithms using explicitly constructed parameters, a separate question is whether training can recover such algorithms from data.

\citet{gatmiry2024learn} investigate this question in the context of noiseless Gaussian in-context linear regression. For a restricted linear-attention model, they show that optima of the population loss correspond to performing multiple steps of preconditioned gradient descent during the forward pass. They further analyze whether gradient flow can recover such solutions under isotropic Gaussian data and additional assumptions on the model parameterization and sample size.

Similarly, \citet{wu2026power} study whether a normalized linear-attention model can learn an iterative algorithm for principal-component prediction. They show that, under population gradient descent, the model's outputs converge to those of the power method. This guarantee relies on rotationally invariant data, unit-norm input columns, an integrability condition, and a specified symmetric initialization. The result establishes convergence of model outputs but does not guarantee convergence of all model parameters to finite values.

Both studies provide evidence that iterative algorithms can emerge through optimization rather than being explicitly encoded in model parameters. However, their guarantees are restricted to specific linear-attention models, data distributions, and optimization settings, and do not directly extend to practical softmax Transformers.

\subsection{Empirical Implications of Theoretical Results}
\label{app:theory_probes}

The preceding theoretical results motivate empirical evaluations that isolate the contributions of different computational resources and examine whether trained models recover the iterative algorithms identified by theory. We propose four evaluations focusing on recurrence depth, workspace, numerical precision, and learned update rules.

\paragraph{Function approximation.}
The approximation results of \citet{xu2024expressive} show that looped Transformers can achieve tighter approximation bounds with more recurrent iterations. However, the constructed parameters may differ across recurrence depths. This motivates two empirical evaluations. First, models can be trained at different recurrence depths using the same training data to examine how training depth affects approximation error and training cost. Second, each trained model can be evaluated at different inference depths to determine whether additional iterations improve approximation accuracy without further training.

\paragraph{Iterative computation.}
The chain-of-thought simulation of \citet{saunshi2025reasoning} uses both additional recurrent iterations and workspace positions. Similarly, the complexity characterization of \citet{merrill2026exact} highlights the complementary roles of recurrent depth and workspace. These findings motivate independently varying the recurrence count and the number of workspace positions on tasks with known reasoning depth. Measuring accuracy and computational cost across these configurations can reveal whether additional iterations or workspace are more beneficial and how the two resources interact.

\paragraph{Formal-language recognition.}
\citet{merrill2026little} construct looped Transformers that recognize regular languages across input lengths using fixed parameters, growing recurrence depth, and growing numerical precision. This motivates examining whether trained models exhibit similar behavior. A looped Transformer can be trained to recognize a regular language using strings of bounded length and then evaluated on longer strings. By independently varying the recurrence count and numerical precision at inference, one can assess how each resource contributes to length generalization.

\paragraph{Learning iterative algorithms.}
\citet{gatmiry2024learn} show that optimal solutions of the population loss for Gaussian in-context linear regression correspond to preconditioned gradient descent. This motivates examining whether trained looped Transformers perform similar updates during inference. Specifically, the predictions at each recurrent iteration can be compared with those obtained after each step of preconditioned gradient descent. Varying the feature covariance can further assess whether the learned update rule generalizes across input distributions. Comparing models with linear and softmax attention can also reveal whether similar iterative behavior emerges beyond the theoretical setting.

\section{Empirical Evidence and Comparison Criteria}
\label{app:evidence}

\begin{table*}[t]
\vspace{-0.9em}
\centering
\small
\setlength{\tabcolsep}{4pt}
\renewcommand{\arraystretch}{1.08}
\begin{tabularx}{\textwidth}{@{}
>{\raggedright\arraybackslash}p{0.16\textwidth}
YYY@{}}
\toprule
\textbf{Study} & \textbf{Experimental Controls} &
\textbf{Evaluation Metrics} & \textbf{Scope and Limitations} \\
\midrule

Huginn~\citep{geiping2026scaling}
& Varies inference depth while holding the checkpoint and evaluation task fixed.
& Normalized HellaSwag accuracy at fixed depths; throughput from a separate decoding benchmark.
& Supports task-specific depth selection. Generalization to other tasks or adaptive policies requires further evaluation. \\

\addlinespace
Synthetic multi-hop reasoning~\citep{kohli2026loopthink}
& Controls the training-hop range and independently varies inference depth and test hop count.
& Accuracy across reasoning-hop counts and recurrence budgets on a synthetic task.
& Reasoning difficulty is controlled through hop count. Generalization to natural-language reasoning remains unestablished by this comparison. \\

\addlinespace
Looped and untied stacks~\citep{saunshi2025reasoning}
& Matches training tokens and executed depth while varying parameter sharing and parameter count.
& Validation perplexity and average accuracy across six mathematical word-problem tasks.
& Demonstrates metric-dependent performance under matched executed depth and token exposure. \\

\addlinespace
MoR-2~\citep{bae2026mixture}
& Matches training tokens while jointly introducing token routing and recursion-wise caching relative to fixed recursion.
& Validation negative log-likelihood, few-shot accuracy, and estimated forward-pass FLOPs.
& Evaluates the combined configuration. Isolating individual contributions requires separate routing and caching ablations. \\

\addlinespace
DeepLoop~\citep{li2026deeploop}
& Controls backbone scale, recurrence count, training data, and seed while varying the overall block parameterization.
& Validation cross-entropy; a separate exponent sweep examines scaling sensitivity.
& Supports the evaluated GPT-2-small configuration at seven loops. Robustness across seeds remains unestablished. \\

\addlinespace
Continuous Depth Batching~\citep{schwethelm2026depth}
& Compares scheduling strategies using recorded exit decisions and fixed response lengths.
& Measured throughput relative to full-depth continuous batching on one H100.
& Covers two model families and three workloads but excludes online exit selection and its effect on output quality. \\

\addlinespace
LLoCoT~\citep{graugarcia2026llocot}
& Matches hardware and input prompts while varying the reasoning procedure; outputs and lengths may differ.
& Measured model-execution time and peak allocated memory; estimated FLOPs.
& Evaluates 128 coding prompts at batch size one with three runs per prompt. Quality on timed outputs and concurrent execution require further evaluation. \\

\bottomrule
\end{tabularx}
\caption{Comparability analysis of the seven studies in Table~\ref{tab:survey_synthesis}. Each row assesses the experimental controls, evaluation metrics, and limitations of the selected within-study comparison.}
\label{tab:comparability_audit}
\vspace{-0.5em}
\end{table*}

This appendix complements Section~\ref{sec:capabilities} by examining how experimental controls and evaluation metrics affect the interpretation of empirical results. Table~\ref{tab:comparability_audit} analyzes the comparisons in Table~\ref{tab:survey_synthesis} in terms of the factors varied, the quantities measured or estimated, and the scope of the evidence. We also provide an evaluation protocol for deployment.

\subsection{Comparison Controls}
\label{app:evidence_controls}

Empirical comparisons address different questions depending on which resources are held fixed. Matching parameter count, training tokens, training FLOPs, or inference FLOPs does not imply that other resources are matched. Table~\ref{tab:evidence_controls} summarizes these comparison settings and their reporting requirements.
Parameter accounting should specify whether embeddings, non-recurrent boundary modules, and iteration-specific parameters are included. For sparse models, stored parameters and active computation should be reported separately. Training-cost estimates should clarify whether backward passes, teacher computation, and auxiliary computation are included, as well as distinguish pretraining, conversion, and post-training costs. Unreported quantities should not be assumed to be matched.

\subsection{Reasoning Capabilities and Generalization}
\label{app:evidence_capabilities}

Fixed-checkpoint evaluations examine what additional inference iterations achieve without changing model parameters, whereas comparisons of separately trained models examine how the choice of recurrence depth during training affects performance. Table~\ref{tab:evidence_capabilities} summarizes representative findings and their limitations.

\paragraph{Inference depth and depth extrapolation.}
Increasing inference depth within a fixed checkpoint can improve prediction quality, as illustrated by the HellaSwag results of \citet{geiping2026scaling}. However, such improvements demonstrate recurrence-depth extrapolation only beyond the depths encountered during training (Section~\ref{subsec:training_depth}). An inference depth above a verified training-depth cap satisfies this condition, whereas depths at or below the cap require knowledge of realized training depths. The mean training depth alone is insufficient.

\paragraph{Reasoning complexity versus recurrence depth.}
Generalization to problems requiring more reasoning steps is distinct from generalization to unseen recurrence depths. \citet{kohli2026loopthink} investigate this distinction through controlled compositional tasks. Table~\ref{tab:survey_synthesis} reports their four-layer model trained with $R_{\rm train}=8$ on problems requiring up to 12 reasoning hops. By holding the training-hop range fixed while varying training recurrence, and separately varying inference depth, their experiments distinguish the effects of reasoning complexity and recurrent computation. Accordingly, evaluations should report $(H_{\rm train},R_{\rm train},H_{\rm test},R_{\rm test})$, where $H$ denotes reasoning hops and $R$ denotes recurrence depth or its distribution.

\paragraph{Comparing models trained at different recurrence depths.}
ETD separately mid-trains configurations that repeat selected internal layers and finds that the benefits of recurrence vary across tasks~\citep{koishekenov2025etd}. Similarly, among the separately trained LoopCoder-v2 variants, the two-loop configuration achieves the highest average benchmark performance, although individual benchmarks exhibit different trends~\citep{yang2026loopcoderv2}. These comparisons inform the choice of recurrence depth during training rather than depth extrapolation at inference. Moreover, the parallel-loop architecture and cross-loop positional offsets of LoopCoder-v2 limit the applicability of its findings to other recurrence structures.

\paragraph{Candidate quality and output selection.}
Evaluating recurrent computation requires distinguishing the quality of intermediate predictions from the ability to select them. \citet{fan2024looped} investigate task-derived iteration counts based on input structure and maximum-confidence selection over a bounded recurrent trajectory. The latter does not necessarily reduce computation because selection may occur after all candidate states have been computed. \citet{kuo2026stabilizing} measure oracle-over-iterations accuracy, defined as the proportion of examples for which at least one evaluated recurrence depth yields the correct answer. Comparing oracle and policy-selected accuracy helps distinguish limitations of candidate quality from those of output selection. Computational savings, however, depend on the computation actually executed rather than the selected prediction depth (Appendix~\ref{app:inference_accounting}).

\begin{table*}[t]
\vspace{-0.5em}
\centering
\small
\setlength{\tabcolsep}{4pt}
\renewcommand{\arraystretch}{1.1}
\begin{tabularx}{\textwidth}{@{}
>{\raggedright\arraybackslash}p{0.18\textwidth}
>{\raggedright\arraybackslash}X
>{\raggedright\arraybackslash}X@{}}
\toprule
\textbf{Comparison Basis} &
\textbf{Evaluation Objective} &
\textbf{Additional Reporting Requirements} \\
\midrule

Stored parameters
& Compare performance under a fixed parameter budget.
& Executed depth, active experts, training compute, and inclusion of embeddings and iteration-specific parameters. \\

\addlinespace

Training tokens
& Compare performance at matched token counts.
& Data composition, compute per token, auxiliary computation, and gradient propagation depth. \\

\addlinespace

Training FLOPs
& Compare performance under a fixed training-compute budget.
& Token counts, training data, FLOP accounting, and inherited pretraining and teacher costs. \\

\addlinespace

Inference FLOPs
& Compare prediction quality at matched inference compute.
& Attention context, readout and stopping overhead, generation length, and training budgets. \\

\addlinespace

Measured runtime or memory
& Compare execution efficiency under a common workload.
& Hardware, precision, kernels, batch size, sequence lengths, concurrency, memory components, and included execution stages. \\

\bottomrule
\end{tabularx}
\caption{Comparison bases, evaluation objectives, and reporting requirements.
The final row concerns measured execution outcomes rather than matched resource budgets.}
\label{tab:evidence_controls}
\vspace{-0.5em}
\end{table*}

\subsection{Parameter and Compute Comparisons}
\label{app:evidence_compute}

Beyond examining the capabilities of recurrent computation, we compare its effectiveness against untied computation under different resource constraints. We discuss how these comparisons depend on evaluation metrics, resource budgets, and architectural choices.

\paragraph{Task-dependent performance.}
The comparison of a 12-layer block applied twice with a 24-layer untied Transformer in \citet{saunshi2025reasoning} controls executed depth and training data (250B tokens from the Pile~\citep{gao2020pile}). As discussed in Section~\ref{subsec:empirical_capabilities}, the looped model improves mathematical reasoning accuracy but exhibits higher validation perplexity.

Benchmark-dependent effects persist across model sizes. In the zero-shot evaluation of \citet{liu2024mobilellm}, immediate layer sharing improves HellaSwag accuracy by 0.6 and 1.0 percentage points at 125M and 350M parameters, respectively, but reduces WinoGrande accuracy by 1.0 and 0.4 points. These consistent trends highlight the importance of task-specific evaluation.

\paragraph{Matched tokens versus matched training compute.}
\citet{bae2026mixture} evaluate a two-recursion MoR configuration with expert-choice routing, middle-cycle sharing, and recursion-wise caching. As summarized in Table~\ref{tab:evidence_mor}, MoR uses fewer FLOPs at matched token counts and processes more tokens at matched FLOPs. The reported FLOPs follow the authors' forward-pass accounting rather than total training computation. Moreover, because routing and caching also change, the improvements cannot be attributed to parameter sharing alone.

MoR is trained on a deduplicated subset of FineWeb-Edu~\citep{penedo2024fineweb,bae2026mixture}. Its few-shot accuracy aggregate covers six benchmark columns: LAMBADA, HellaSwag, PIQA, WinoGrande, ARC, and MMLU~\citep{paperno2016lambada,zellers2019hellaswag,bisk2020piqa,sakaguchi2020winogrande,clark2018arc,hendrycks2021mmlu}. ARC combines ARC-Easy and ARC-Challenge, while MMLU uses continuation-based evaluation. These aggregation and evaluation conventions should be considered when interpreting cross-study comparisons.

\paragraph{Training versus inference compute.}
Matching inference compute does not necessarily control training compute. For example, the from-scratch LoopLM and dense baselines of \citet{marchenko2026closing} use the same training data and token budgets. However, the 3.9B dense baseline matches the inference computation of three LoopLM recurrences, whereas LoopLM is trained through four.

\begin{table*}[t]
\vspace{-0.4em}
\centering
\small
\setlength{\tabcolsep}{4pt}
\renewcommand{\arraystretch}{1.1}
\begin{tabularx}{\textwidth}{@{}
>{\raggedright\arraybackslash}p{0.20\textwidth}
>{\raggedright\arraybackslash}X
>{\raggedright\arraybackslash}X@{}}
\toprule
\textbf{Study} &
\textbf{Empirical Findings} &
\textbf{Scope and Limitations} \\
\midrule

\citet{yang2024looped}
& Substantial parameter savings on synthetic in-context prediction tasks; linear regression experiments also examine variable inference depths.
& Evidence is limited to synthetic tasks; parameter savings do not imply runtime savings. \\

\addlinespace

\citet{saunshi2025reasoning}
& Language models trained on 250B Pile tokens exhibit task-dependent performance relative to untied models.
& Separately trained configurations do not establish fixed-checkpoint depth extrapolation. \\

\addlinespace

\citet{fan2024looped}
& Algorithmic length generalization with task-derived iteration counts; additional experiments examine confidence-based output selection.
& Task-derived depths use input structure rather than reference answers; confidence-based selection may require executing the full trajectory. \\

\addlinespace

\citet{geiping2026scaling}
& Additional inference iterations improve performance across several benchmarks within the same Huginn checkpoint.
& Exceeding the mean training depth does not necessarily establish depth extrapolation. \\

\addlinespace

\citet{jeddi2026loopformer}
& A single model supports multiple budget-conditioned trajectories within the maximum training depth.
& Changing the recurrence budget alters the conditioning; shorter trajectories need not be prefixes of longer ones. \\

\addlinespace

\citet{zhu2026ouro}
& Controlled synthetic probes distinguish factual storage from the manipulation and composition of stored information.
& Probe-specific findings do not establish a general distinction between memory and reasoning. \\

\addlinespace

\citet{kuo2026stabilizing}
& Training-depth schedules affect extrapolation stability; additional experiments compare oracle and policy-selected accuracy on Unique Set.
& Prediction accuracy and trajectory stability do not directly establish execution savings. \\

\addlinespace

ETD~\citep{koishekenov2025etd}
& Repeating selected internal layers improves reasoning performance, with the preferred recurrence count varying across tasks.
& Separately mid-trained configurations establish neither fixed-checkpoint depth extrapolation nor gains at matched training FLOPs. \\

\addlinespace

LoopCoder-v2~\citep{yang2026loopcoderv2}
& The two-loop configuration achieves the highest benchmark average among the evaluated variants.
& Variants are trained separately; the parallel-loop design limits generalization to other recurrent architectures. \\

\addlinespace

\citet{kohli2026loopthink}
& Controlled compositional experiments vary reasoning hops and recurrence depths during training and inference, including matched-hop training comparisons.
& Reasoning-hop extrapolation differs from recurrence-depth extrapolation. Synthetic findings do not establish general scaling behavior in NLP. \\

\bottomrule
\end{tabularx}
\caption{Representative empirical findings on looped Transformer capabilities and generalization, with their scope and limitations. Differences in experimental settings limit direct cross-study comparisons.}
\label{tab:evidence_capabilities}
\vspace{-0.9em}
\end{table*}

\paragraph{Increasing depth versus replacing distinct layers.}
RINS~\citep{alabdulmohsin2025rins} increases computational depth at a fixed parameter count by repeating selected blocks. Because repetition increases computation per token, recurrent configurations receive fewer training tokens to match the reported forward-pass FLOP budget. Selected RINS configurations nevertheless achieve lower language-modeling loss.

In contrast, \citet{schwethelm2026recurrence} replace distinct layers with shared recurrent applications, reducing parameter count while keeping effective depth fixed. At a given width, per-token computation remains approximately constant. By varying model width and training-token allocation, they estimate a compute-optimal frontier for each recurrence count and find higher validation loss for the baseline looped configurations.

These findings suggest that the benefits of increasing depth through parameter reuse do not necessarily extend to replacing distinct layers with shared ones. Differences in architecture and training protocol prevent attributing the contrasting results to a single factor.

\paragraph{Scaling contribution of recurrence.}
Classical language-model scaling laws characterize the relationships among model size, training data, and computational cost~\citep{kaplan2020scaling,hoffmann2022computeoptimal}, but do not explicitly distinguish recurrent computation from additional distinct parameters. To investigate this distinction, \citet{schwethelm2026recurrence} conduct 116 iso-depth pretraining runs with $R\in\{1,2,4,8\}$ and introduce an effective parameter proxy in their fitted loss model:
\begin{equation}
P_{\mathrm{eq}}
=
P_{\mathrm{once}}
+
R^{\varphi}P_{\mathrm{rec}},
\label{eq:evidence_recurrence_proxy}
\end{equation}
where $P_{\mathrm{once}}$ and $P_{\mathrm{rec}}$ denote the non-embedding parameter counts outside and inside the recurrent core, respectively. The baseline fit yields $\varphi\approx0.46$, so four applications of the recurrent core contribute approximately $1.9P_{\mathrm{rec}}$ to the effective parameter proxy. This suggests diminishing returns from additional recurrences within the fitted loss model. However, $P_{\mathrm{eq}}$ is a fitted proxy specific to the evaluated architectures and training regime, rather than an actual parameter count or a universal measure of model capacity.

\begin{table*}[t]
\vspace{-0.5em}
\centering
\small
\setlength{\tabcolsep}{5pt}
\begin{tabular}{@{}lccccc@{}}
\toprule
\textbf{Configuration} & \textbf{Non-emb. Params}
& \textbf{FLOPs ($10^{18}$)} & \textbf{Training Tokens}
& \textbf{NLL $\downarrow$} & \textbf{Avg. Acc. (\%) $\uparrow$} \\
\midrule
Vanilla                   & 315M & 16.5 & 20B & 2.7824 & 42.3 \\
Fixed recursion, $R=2$    & 167M & 16.5 & 20B & 2.8079 & 42.6 \\
MoR-2 (matched FLOPs)     & 167M & 16.5 & 27B & 2.7511 & 43.1 \\
MoR-2 (matched tokens)    & 167M & 12.3 & 20B & 2.7749 & 42.9 \\
\bottomrule
\end{tabular}
\caption{Training comparisons under matched FLOPs or matched tokens~\citep{bae2026mixture}. MoR-2 uses expert-choice routing, middle-cycle sharing, and recursion-wise caching. NLL denotes validation negative log-likelihood, and accuracy is averaged across six benchmark columns. Reported FLOPs follow the source's forward-pass accounting and exclude backward computation.}
\label{tab:evidence_mor}
\vspace{-0.5em}
\end{table*}

\paragraph{Sparse architectures.}
Sparse architectures distinguish stored capacity from active computation. SMELT compares looped and untied MoE architectures with closely matched per-token FLOPs, non-embedding parameter counts, and KV-cache budgets~\citep{wang2026smelt}. Despite its favorable loss--compute trade-off, the comparison does not isolate the effect of parameter sharing.

Performance also depends on the evaluation metric. \citet{lee2026sparse} find that untied MoE outperforms looped MoE in full-depth test-loss scaling. However, at a reported training budget of $10^{18}$ FLOPs, looped MoE achieves a higher Core 9 downstream average ($39.6$ vs.\ $36.4$), computed under the study's OLMES-based evaluation protocol~\citep{gu2025olmes}. As in dense architectures, improvements in downstream accuracy do not necessarily translate into lower language-modeling loss.

\paragraph{Comparing parameterizations.}
DeepLoop's GPT-2-small comparison at $R=7$ controls architecture scale, training data (50B FineWeb-Edu tokens), optimizer, and seed while varying the residual parameterization~\citep{li2026deeploop}. Its improvement over the pre-normalized baseline therefore evaluates the overall parameterization rather than an individual scaling exponent. Moreover, the single-seed comparison does not establish robustness across runs. Appendix~\ref{app:training_parameterization} discusses the separate scaling-exponent ablation.

\subsection{Interpreting Systems Measurements}
\label{app:evidence_systems}

Parameter and FLOP comparisons alone do not determine runtime efficiency. Systems evaluations should distinguish estimated arithmetic costs, measurements of individual execution components, and end-to-end serving performance.

\paragraph{Arithmetic cost versus measured execution.}
Arithmetic estimates do not capture memory traffic, hardware utilization, or scheduling overhead. FlashAttention demonstrates how I/O-aware implementations can improve execution efficiency without changing the underlying attention computation~\citep{dao2022flashattention}. Conversely, MoEUT reports slower training than its dense FlashAttention baseline despite lower arithmetic cost~\citep{csordas2024moeut}. The early-exit savings reported by \citet{lee2026sparse} similarly concern estimated inference arithmetic rather than measured throughput.

\paragraph{Generation and serving measurements.}
Recursive Transformers estimate throughput by combining measured Transformer-block execution times with oracle exit trajectories based on agreement with final-layer predictions~\citep{bae2025relaxed}. This captures potential early-exit benefits rather than the performance of an online stopping policy.

MoR reports up to $2.06\times$ generation throughput in a four-recursion experiment, using different batch sizes for MoR and its effective-depth-matched dense baseline~\citep{bae2026mixture}. However, the measurements exclude KV-cache insertion and updates. This experiment is also distinct from the two-recursion training comparisons discussed in Appendix~\ref{app:evidence_compute}.

Continuous Depth Batching evaluates scheduling efficiency by replaying recorded exit decisions and fixing response lengths to those of reference outputs~\citep{schwethelm2026depth}. This protocol, described in Appendix~C of their paper, isolates scheduling performance but excludes the cost of selecting exits during generation.

\begin{table*}[t]
\vspace{-0.5em}
\centering
\small
\setlength{\tabcolsep}{4pt}
\renewcommand{\arraystretch}{1.1}
\begin{tabularx}{\textwidth}{@{}
>{\raggedright\arraybackslash}p{0.23\textwidth}YY@{}}
\toprule
\textbf{Empirical Insight} &
\textbf{Supporting Evidence} &
\textbf{Further Evaluation} \\
\midrule

Language-modeling loss and downstream accuracy can favor different configurations.
& Dense-model comparisons, fixed-recursion baselines, and sparse-model evaluations~\citep{saunshi2025reasoning,bae2026mixture,lee2026sparse}.
& Consistency of model rankings across metrics, statistical uncertainty, and performance on deployment tasks. \\

\addlinespace

Additional inference iterations can improve performance, but gains vary across settings.
& Huginn depth sweeps and controlled experiments on compositional reasoning and overthinking~\citep{geiping2026scaling,kohli2026loopthink}.
& Effects of training-depth exposure, input complexity, and adaptive depth selection. \\

\addlinespace

Effective output selection does not guarantee computational savings.
& Oracle--policy accuracy gaps, differences between selected and executed depths, and replayed-exit scheduling~\citep{kuo2026stabilizing,vendrell2026memory,schwethelm2026depth}.
& Output selection under identical candidate sets, followed by end-to-end evaluation of online policies. \\

\addlinespace

Lower FLOPs do not necessarily imply lower memory usage or latency.
& MoEUT implementation costs, LT2 mixer designs, and LLoCoT serial-depth trade-offs~\citep{csordas2024moeut,deng2026lt2,graugarcia2026llocot}.
& Comparisons at comparable prediction quality with explicit runtime and memory accounting. \\

\bottomrule
\end{tabularx}
\caption{Synthesis of empirical evidence and remaining evaluation needs. The findings are specific to the evaluated settings rather than universal model rankings. The final column identifies proposed evaluations.}
\label{tab:claim_evidence}
\vspace{-0.5em}
\end{table*}

\paragraph{Verification-based execution.}
Efficiency measurements must also account for verification when intermediate predictions are used for drafting rather than directly accepted. LoopSpec uses shallower recurrent states to draft tokens and deeper computation to verify them, together with pipelined execution~\citep{cho2026loopspec}. Its verification procedure is designed to preserve the target decoding behavior, while efficiency depends on draft acceptance rates, verification overhead, and workload characteristics.

\paragraph{Serial computation, arithmetic, and memory.}
Reducing serial computation can improve latency despite increasing arithmetic cost and memory consumption. LLoCoT evaluates efficiency on one H100 80\,GB using BF16 precision, batch size one, and 128 fixed coding prompts, with three runs after ten warmups~\citep{graugarcia2026llocot}. Compared with Reasoning SFT, LLoCoT reduces reasoning time from $2.47$ to $0.06$\,s and model-execution time from $15.16$ to $11.47$\,s, while increasing peak memory from $3.3$ to $6.3$\,GiB and estimated FLOPs from $1.7\times10^{12}$ to $2.8\times10^{12}$.

The timings exclude tokenization, sampling, and evaluation overhead. Moreover, outputs and response lengths can differ, and the separate quality benchmark does not establish equal prediction quality on the timing prompts.

\subsection{Design and Evaluation Protocol}
\label{app:design_evaluation}

Table~\ref{tab:claim_evidence} synthesizes the preceding findings and identifies remaining evaluation needs. We propose an evaluation protocol for selecting looped architectures under the deployment constraints in Section~\ref{subsec:design_guide}, distinguishing attainable prediction quality, adaptive policy effectiveness, and implementation efficiency.

\paragraph{Specify the deployment objective.}
Identify the quality metric, acceptable quality loss, workload distribution, and limiting resource. Select baselines and account for resources according to Table~\ref{tab:evidence_controls}. For stochastic evaluations, report uncertainty and define any non-inferiority tolerance before testing.

\paragraph{Establish a fixed-depth reference.}
Evaluate the same checkpoint over a range of recurrence depths under a common workload, reporting prediction quality, executed updates, FLOPs, and measured latency separately. Record training-depth exposure, input lengths, and reasoning complexity. For budget-conditioned architectures such as LoopFormer~\citep{jeddi2026loopformer}, changing the recurrence budget may alter the entire trajectory rather than simply extend or truncate it. In addition to aggregate accuracy, report how often later iterations correct earlier errors or replace correct predictions with incorrect ones.

\paragraph{Evaluate candidate quality and selection.}
For tasks with a defined correctness criterion, compare oracle-over-candidates accuracy with policy-selected accuracy~\citep{kuo2026stabilizing}. Report the candidate depths, information available to the policy, prediction quality, and executed-depth distributions. Use reference answers only for offline oracle evaluation, select stopping thresholds on validation data, and evaluate on held-out data.

Oracle accuracy bounds only selection among the evaluated candidates. In autoregressive generation, changing an earlier prediction may alter subsequent tokens and cache contents, while adaptive routing may change the attention context. Thus, an oracle computed from a recorded trajectory does not necessarily bound the performance of alternative online policies.

\paragraph{Measure the deployed policy.}
Compare the actual stopping, routing, and caching policies with the fixed-depth reference under identical task and hardware conditions. Account for intermediate readouts, exit decisions, routing, cache operations, and synchronization. Report prefill time, time to first output, decoding latency, total response latency, throughput, and output-length distributions, specifying any excluded operations.

Report memory usage separately for weights, KV caches or recurrent states, temporary buffers, and peak allocation. Specify hardware, precision, kernels, device count, batch size, and concurrency. Profile representative sequence lengths and workloads to identify bottlenecks in arithmetic, memory traffic, or scheduling~\citep{schwethelm2026depth}.

\begin{table*}[t]
\vspace{-0.5em}
\centering
\small
\setlength{\tabcolsep}{5pt}
\renewcommand{\arraystretch}{1.12}
\begin{tabularx}{\textwidth}{@{}l
>{\raggedright\arraybackslash}p{0.47\textwidth}
>{\raggedright\arraybackslash}X@{}}
\toprule
\textbf{ID} &
\textbf{Search Query} &
\textbf{Publication-Site Restrictions} \\
\midrule

Q1 &
\texttt{"looped transformer" "parameter sharing"} &
None \\

Q2 &
\texttt{"recurrent-depth" transformer} &
None \\

Q3 &
\texttt{"recursive transformer" "language"} &
None \\

Q4 &
\texttt{"universal transformer" "halting"} &
None \\

Q5 &
\texttt{"looped transformers"} &
arXiv; PMLR; NeurIPS proceedings \\

Q6 &
\texttt{"recursive transformer"} &
ACL Anthology; ICLR proceedings; OpenReview \\

Q7 &
\texttt{"universal transformer" "adaptive"} &
ACL Anthology; OpenReview \\

Q8 &
\texttt{"recurrent" "parameter sharing" "transformer"} &
ACL Anthology; NeurIPS proceedings \\

\bottomrule
\end{tabularx}
\caption{Search queries and publication-site restrictions for the supplementary literature search on 12 October 2026. Quotation marks indicate exact-phrase queries. Site restrictions do not imply exhaustive source coverage.}
\label{tab:review_search_queries}
\vspace{-0.5em}
\end{table*}

\paragraph{Account for training and serving costs.}
Additional training costs may be offset by serving savings. Let $T_j$ denote the training cost of configuration $j$ and $s_j$ its mean serving cost per request under a fixed workload and load regime. Using a common additive unit, such as money or energy, the total training and serving cost for $N$ requests is
\begin{equation}
C_j(N)=T_j+Ns_j.
\label{eq:lifecycle_cost}
\end{equation}

Let $L$ and $U$ denote a looped candidate and its untied baseline, respectively. If the looped candidate incurs a higher training cost ($\Delta T=T_L-T_U>0$) but a lower serving cost ($s_L<s_U$), the break-even request volume is
\begin{equation}
N_*=\frac{\Delta T}{s_U-s_L}.
\label{eq:lifecycle_breakeven}
\end{equation}

Under this model, the looped candidate has a lower total cost for $N>N_*$. Without serving savings, the additional training cost cannot be recovered.

Training costs should follow the accounting requirements in Appendix~\ref{app:evidence_controls}. Shared pretraining costs may cancel when comparing adaptations from a common checkpoint but remain part of model provenance. Training-free conversion eliminates additional parameter optimization but not inherited pretraining or additional inference computation. Compare total costs only between configurations satisfying the application's quality, memory, and latency requirements, and account for uncertainty in measured costs and expected request volume.

\paragraph{Identify the next evaluation.}
If few correct predictions are available across recurrent depths, investigate training objectives or architectural changes. If correct candidates are available but frequently missed, investigate output selection and stopping policies. If adaptive computation reduces executed updates without improving runtime, investigate caching, scheduling, and hardware utilization.

\section{Review Procedure and Evidence Selection}
\label{app:review_procedure}

\subsection{Supplementary Literature Search}
\label{app:review_search}

We conducted a targeted narrative review rather than an exhaustive systematic review. The initial bibliography was assembled during manuscript drafting, with language-model suggestions used to identify potentially relevant studies. All descriptions of architectures, training procedures, empirical findings, and theoretical results were based on the corresponding primary papers.

To assess the coverage of the initial bibliography and identify additional relevant studies, we conducted a supplementary literature search on 12 October 2026 using the eight queries listed in Table~\ref{tab:review_search_queries}. The search covered web-indexed records subject to the publication-site restrictions specified in the table, with no lower bound on the publication year. We also examined selected references from the study introducing Huginn, a recurrent-depth language model~\citep{geiping2026scaling}, and from \textit{A Survey on Latent Reasoning}~\citep{zhu2025latentsurvey}.

Studies identified through the supplementary search and reference checks were considered alongside those in the initial bibliography. We screened titles and abstracts and examined selected primary papers to assess their relevance to the scope defined in Section~\ref{sec:definition_scope}. The selected studies informed our taxonomy, empirical comparisons, theoretical discussion, and supporting technical details. We do not report separate inclusion counts for each literature identification strategy.

During manuscript revision, we further included the survey by \citet{azizi2026looped}, released on 15 October 2026, to compare the scopes of existing surveys, and MobileLLM~\citep{liu2024mobilellm} to examine empirical evidence on on-device execution.

\subsection{Evidence Selection and Review Limitations}
\label{app:review_evidence}
\label{app:review_limits}

We distinguish studies that evaluate configurations within our scope from those cited for theoretical, algorithmic, or technical background. Since citations serve different purposes and individual studies may evaluate multiple configurations, the number of cited studies does not directly reflect the number of configurations reviewed.

For model-specific claims, we relied on primary papers and treated preprint and conference versions of the same study as a single reference. For empirical studies, we recorded architectural choices, training depths and objectives, inference depths and cache policies, along with the reported baselines, evaluation metrics, and resource controls. For theoretical studies, we recorded assumptions concerning inputs, weights, numerical precision, and computational resources.

We selected representative studies to illustrate design choices and their associated trade-offs, including cases in which reported gains disappear when evaluated using alternative metrics or baselines. However, our reliance on web-indexed searches and selective reference checks may have resulted in the omission of relevant studies. Appendix~\ref{app:evidence} provides a detailed analysis of the comparability of empirical results and the conclusions supported by the available evidence.

\end{document}